\pdfoutput=1

\documentclass[11pt]{article}

\usepackage[final]{acl}

\usepackage{times}
\usepackage{latexsym}

\usepackage[T1]{fontenc}

\usepackage[utf8]{inputenc}

\usepackage{microtype}

\usepackage{inconsolata}

\usepackage{graphicx}

\usepackage{graphicx}
\usepackage{amsmath}

\usepackage{algorithm}
\usepackage{algpseudocode}

\usepackage[greek,english]{babel}

\usepackage{adjustbox}
\usepackage{multirow}
\usepackage{xcolor}    
\definecolor{darkgreen}{rgb}{0.0, 0.5, 0.0} 
\usepackage{booktabs}
\usepackage{afterpage}

\usepackage{listings}

\newcommand{\daggerfootnote}[1]{%
  \begingroup
  \renewcommand{\thefootnote}{\fnsymbol{footnote}}%
  \setcounter{footnote}{2}
  \footnotetext{#1}%
  \setcounter{footnote}{0}
  \renewcommand{\thefootnote}{\arabic{footnote}}%
  \endgroup
}

\newlength{\myalgindent}
\algnewcommand\LeftComment[2]{%
  \hspace{#1\myalgindent}$\triangleright$ \text{#2}%
}

\newcommand\Algphase[1]{%
  \vspace*{-.1\baselineskip}\Statex\hspace*{0pt}\rule{\linewidth}{0.4pt}%
  \Statex\hspace*{\algorithmicindent}\textbf{#1}%
  \vspace*{-.1\baselineskip}\Statex\hspace*{0pt}\rule{\linewidth}{0.4pt}%
}

\title{Learning to Align: Addressing Character Frequency Distribution Shifts in Handwritten Text Recognition}

\author{
  \textbf{Panagiotis Kaliosis\textsuperscript{1,2,\textdagger}} \hspace{0.02cm}
  \textbf{John Pavlopoulos\textsuperscript{2,3}}
\\
\\
  \textsuperscript{1}Stony Brook University, New York, USA \\
  \textsuperscript{2}Archimedes, Athena Research Center, Greece \\
  \textsuperscript{3}Athens University of Economics and Business, Greece \\
  \normalsize\texttt{pkaliosis@cs.stonybrook.edu, annis@aueb.gr}
}

\begin{document}
\maketitle
\begin{abstract}
Handwritten text recognition aims to convert visual input into machine-readable text, and it remains challenging due to the evolving and context-dependent nature of handwriting. Character sets change over time, and character frequency distributions shift across historical periods or regions, often causing models trained on broad, heterogeneous corpora to underperform on specific subsets. To tackle this, we propose a novel loss function that incorporates the Wasserstein distance between the character frequency distribution of the predicted text and a target distribution empirically derived from training data. By penalizing divergence from expected distributions, our approach enhances both accuracy and robustness under temporal and contextual intra-dataset shifts. Furthermore, we demonstrate that character distribution alignment can also improve existing models at inference time without requiring retraining by integrating it as a scoring function in a guided decoding scheme. Experimental results across multiple datasets and architectures confirm the effectiveness of our method in boosting generalization and performance. We open source our code at \href{https://github.com/pkaliosis/fada}{this link}.
\end{abstract}

\section{Introduction}
\daggerfootnote{Work done during internship at Archimedes Research Center before joining Stony Brook University.}
Handwritten Text Recognition (HTR) enables the automatic conversion of handwritten input into machine-readable text, supporting critical applications such as historical manuscript digitization, document analysis, and archival research. 
Although often overlooked, systematic shifts in character frequency distributions, caused by linguistic evolution, orthographic reforms, and contextual usage, are an important consideration when building robust HTR systems across diverse time periods or regions \cite{moreno2005frequency}. 
In the transition from Old English to Modern English, for example, grammatical changes, such as the loss of case markings, led to notable alterations in letter distributions \cite{moreno2005frequency}. Character frequency distributions also differ across languages that share the same alphabet, with measurable disparities even among closely related languages \cite{grigas-frequencies-latin}. Beyond language, writing style and genre exert substantial influence on letter usage, which is reflected in systematic differences across news articles, novels, and scientific texts \cite{Zhao2024-frequency-genres}. Specialized domains further introduce deviations: for instance, proprietary prescription drug names diverge markedly from standard English distributions \cite{carico-frequency-drugs}, while authorship attribution studies exploit subtle statistical variations in letter frequencies to differentiate between writers \cite{diurdeva-freq-author1, Merriam-freq-author2, Kjell-freq-author3}. These findings underscore the dynamic nature of character frequency distributions and their dependence on both linguistic and contextual factors.

HTR models face challenges when transcribing historical manuscripts, where orthographic conventions and character usage shift considerably over time \cite{htr-tsochatzidis, Cascianelli2022BoostingMA}. A common mitigation strategy involves training on diverse corpora to increase generalization, but this does not necessarily yield optimal performance \cite{nguyen2022quality}. 
In initial experiments, we observed that models trained on heterogeneous data tend to underperform in specific subsets where character frequency distributions deviate, leading to systematic biases and suboptimal transcription accuracy, a phenomenon that to the best of knowledge has not been thoroughly explored in prior work.

To address these challenges, we propose \texttt{FADA} (Frequency-Aware Distribution Alignment), a novel training framework that aligns a model’s predicted character-level frequency distributions with empirically observed ones. When such distributions are available, \texttt{FADA} enhances robustness to temporal and contextual variation by incorporating a distance-based loss function that penalizes discrepancies between predicted and empirical character frequencies during training. This alignment loss allows the model to learn general task-relevant representations while also reflecting the expected subset-specific statistical patterns. In addition, we introduce a guided decoding algorithm that integrates the frequency-aware objective into the beam search process at inference time. By treating empirical character frequency distributions as soft constraints, our method refines predictions to better match expected patterns, improving transcription accuracy without the need for retraining. \texttt{FADA} is the first framework to explicitly address character-level frequency distribution shifts in text recognition. In contrast to prior work, which largely overlooks intra-dataset variation, \texttt{FADA} introduces both a trainable alignment mechanism and an inference-time correction strategy, providing a comprehensive solution for mitigating distributional shifts. 

Our main contributions are: (\romannumeral 1) We introduce \texttt{FADA}, a novel training framework that explicitly addresses character-level frequency distribution shifts in recognition tasks. (\romannumeral 2) We propose a guided decoding strategy that incorporates empirical frequency priors into beam search, enhancing alignment between predicted and expected character frequency distributions at inference time. (\romannumeral 3) We demonstrate the effectiveness and generality of \texttt{FADA} through extensive experiments on HTR, showing consistent performance gains across multiple datasets, languages, and model architectures.

\section{Related Work}

\textbf{Feature and distribution alignment} have been extensively explored as strategies to mitigate performance degradation caused by distribution shifts in recognition models. Prior work addresses these challenges by enhancing alignment at various stages of the recognition pipeline, including structural feature extraction, semantic representation learning, and distribution matching.

In \textbf{scene text recognition (STR)}, \citet{hu-shape-driven} proposed shape-driven attention to guide models toward higher-quality character features using geometric priors. In \textbf{vision-language models (VLMs)}, distributional alignment has also been explored \cite{Cho2023DistributionAwarePT}. For example, \citet{li-discr-captions} minimized distributional gaps between positive and negative captions to improve image-text alignment by promoting stronger visual grounding.

In \textbf{speech-related tasks}, alignment techniques have been applied to adapt to domain-specific variation. \citet{zhou-multi-genre} investigated distribution alignment for multi-genre speaker recognition, showing that Within-Between Distribution Alignment (WBDA) \cite{hu-wbda} improved robustness but did not fully eliminate genre-specific variability. Moreover, \citet{Hou2021CrossdomainSR} aligned the character-level audio representations between a source and a target domain in an unsupervised manner via Maximum Mean Discrepancy.

Standard decoding strategies like greedy or beam search often lack mechanisms to enforce \textbf{external constraints}, which are crucial in tasks requiring adherence to linguistic structure or statistical properties. To address this, \textbf{guided decoding} methods steer generation based on predefined constraints \cite{wang-gd-survey}, typically grouped as semantic, structural, or lexical \cite{Zhang2022ASO}. Semantic constraints enforce stylistic or topical alignment \cite{yang-gd-style1, ghazvininejad-etal-2017-hafez, pascual-etal-2021-plug-play}, or integrate task-specific priors \cite{kaliosis-etal-2024-data, samprovalaki-2024-aueb, chatzipapadopoulou-2025-aueb}. Structural approaches promote syntactic coherence \cite{bastan-etal-2023-neurostructural, lu-etal-2021-neurologic}, while lexical constraints guide the inclusion or exclusion of specific tokens \cite{anderson-etal-2017-guided, hokamp-liu-2017-lexically, yao2024collie}.

Despite this progress, existing approaches do not explicitly address the impact of character-level frequency distribution shifts, which are particularly prominent in recognition tasks involving historical or domain-specific data. Most alignment methods focus on feature-level or semantic alignment, overlooking statistical priors at the character level that can guide model predictions. \textbf{\texttt{FADA}} fills this gap by introducing a frequency-aware training and decoding framework that directly incorporates character-level distributional knowledge into both learning and inference.

\section{The Proposed \texttt{FADA} Method}
\label{sec:method}

Our proposed method, \texttt{FADA}, introduces a novel training framework that aligns the predicted character-level frequency distribution of the model with an empirical target distribution at each training step, thereby reducing discrepancies caused by temporal and contextual shifts.

For example, if the model predicts a character frequency of $9.3\%$ for the letter ``a'', while the empirical distribution indicates an expected $11.1\%$, \texttt{FADA} encourages the model to adjust its predictions to reduce this gap. To achieve this, we introduce an auxiliary loss term that penalizes divergence between the predicted and empirical character frequency distributions. This alignment loss is jointly optimized with the task-specific objective (e.g., CTC or Cross-Entropy loss), enabling the model to preserve recognition performance while adapting to subset-specific distributional properties. As a result, the model becomes more robust to intra-dataset shifts and better aligned with the statistical patterns present in the data.

\paragraph{Empirical Relative Frequency Distributions}

To capture character frequency variations, we compute empirical relative frequency distributions from the training data. First, we normalize transcriptions by converting all text to lowercase to ensure consistency. Frequencies are calculated by counting character occurrences and normalizing by the total number of characters. Intra-dataset distribution shifts are systematic variations in character frequency distributions across distinct subsets of a corpus. To model such shifts, we compute separate frequency distributions for each temporal or contextual subset of the training set (e.g., distinct centuries/regions).

\paragraph{Character Frequency Analysis}

\begin{figure}[th!]
    \centering
    \includegraphics[width=.5\textwidth, height=7cm]{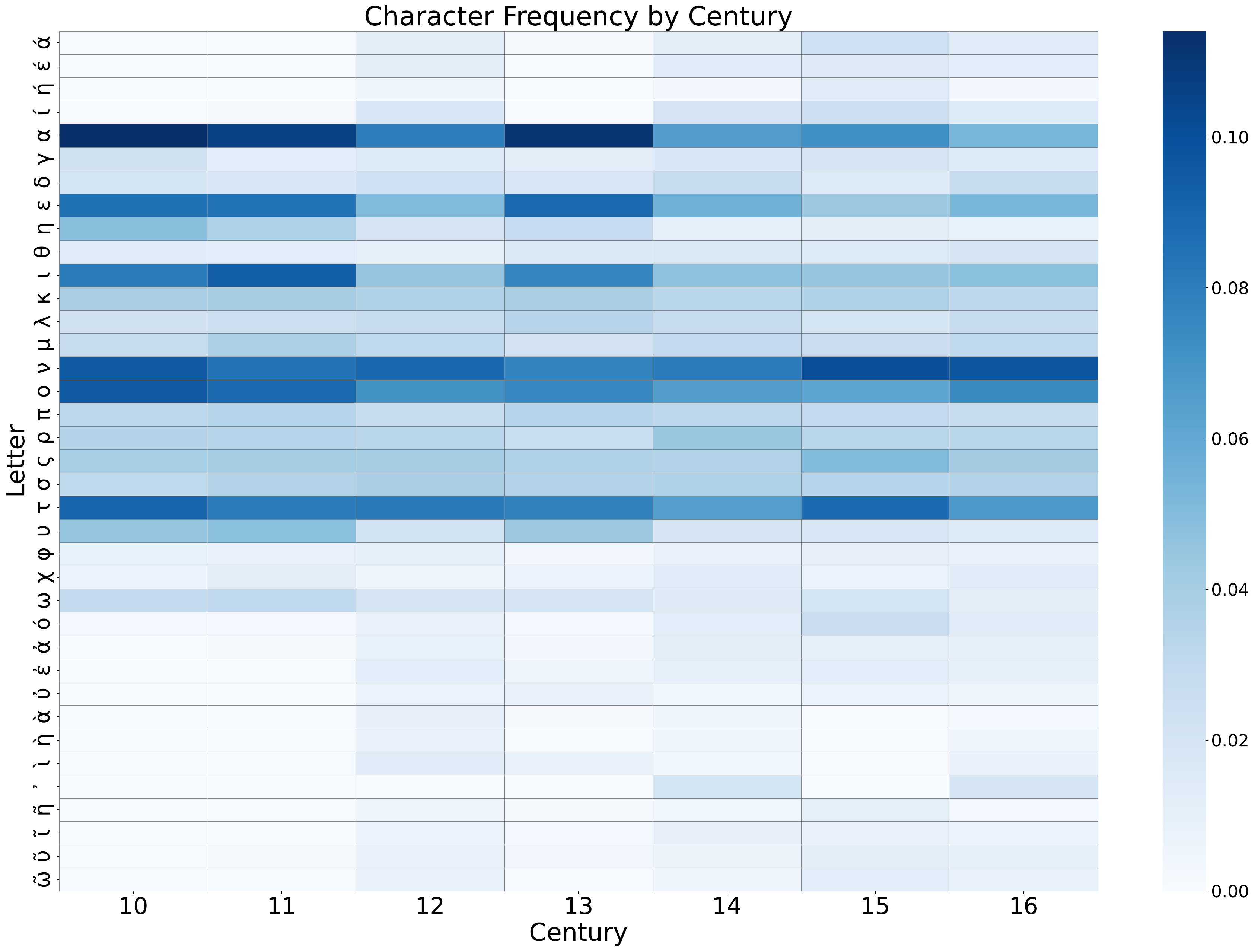}
    \vspace{-3mm}
\caption{Empirical character-level frequency distributions in the HPGT dataset across seven centuries (10th–16th CE). Characters with relative frequency below $0.01$ are omitted for clarity.}
\label{fig:1} 
\end{figure}

Figure~\ref{fig:1} illustrates the character-level relative frequency distributions of the HPGT dataset \cite{platanou-etal-2022-handwritten} over seven centuries, revealing clear distributional shifts in character usage. Certain letters, such as \textgreek{δ}, \textgreek{κ}, and \textgreek{γ}, display stable frequencies over time, suggesting consistent orthographic roles in written Greek. In contrast, others, such as \textgreek{τ} and \textgreek{ε}, exhibit notable fluctuations, suggesting potential shifts in orthographic conventions or variations in dataset composition. Additionally, characters such as ῶ and ἔ, are absent in earlier centuries but emerge in later ones, which may reflect historical linguistic developments or artifacts of dataset annotation and coverage. However, these variations are not solely attributable to linguistic evolution; external factors, such as dataset biases or uneven temporal representation, may also influence the observed distributions. We present a detailed analysis including correlation scores between the character frequency distributions across centuries in Appendix~\ref{sec:appendix-hpgt}.


\paragraph{FADA Training Framework  (\texttt{FADA}\textsubscript{TR})} 

In conventional recognition models, training is typically guided by task-specific objectives such as CTC or CE loss. While these objectives effectively optimize transcription accuracy, they do not account for intra-dataset shifts in character frequency distributions. As a result, models trained on aggregated datasets may develop implicitly biases toward dominant character distributions, leading to suboptimal performance when applied to subsets that exhibit different statistical properties at the character level.

To address this limitation, we introduce an auxiliary \textit{distribution alignment loss} that encourages the model’s predicted character frequency distributions to align with empirical distributions observed in the training data. At each training step, we compute the predicted distribution from the model’s output logits and compare it with a sample-specific empirical distribution derived from the training set (e.g., based on the sample's century or region). The alignment loss penalizes discrepancies between the predicted and empirical distributions, guiding the model toward representations that preserve the expected statistical properties of the corresponding subset. This loss is designed to complement—not replace—the primary recognition objective (CTC or CE), thereby enhancing the model’s robustness to intra-dataset distributional shifts without compromising transcription accuracy.

To formally define this alignment loss, we employ the \textit{Wasserstein distance}, a well-established metric for quantifying discrepancies between probability distributions. Given two probability distributions  $D$  and  $Q$ , the Wasserstein-$p$ ($W_p$)  distance is defined as $W_p(D, Q) = \left( \frac{1}{n} \sum_{i=1}^{n} \| D_{(i)} - Q_{(i)} \|^p \right)^{\frac{1}{p}}$
where $D_{(i)}$ and $Q_{(i)}$ denote the $i$-th elements of the sorted (in ascending order) versions of $D$ and $Q$, respectively, and $n$ is the number of characters.
Since we operate over discrete relative frequency distributions, i.e., one-dimensional empirical probability vectors summing to one, we specifically adopt the second-order Wasserstein distance ($W_2$), where $p=2$:
\begin{equation}
\mathcal{L}_{W_2} = W_2(D, Q) = \left( \frac{1}{n} \sum_{i=1}^{n} \| D_{(i)} - Q_{(i)} \|^2 \right)^{\frac{1}{2}}
\label{eq:1}
\end{equation}

This choice is motivated by the fact that  $W_2$  corresponds to the Euclidean norm, offering a natural and computationally efficient measure of distributional divergence that is sensitive to fine-grained differences. In the one-dimensional case, $W_2$ simplifies to the Root Mean Squared Error (RMSE) between the sorted frequency vectors, making it especially suitable for character-level distributions.
At each training step, the total loss is defined as a convex combination of the primary recognition loss and the distribution alignment loss. The former corresponds to either CTC ($\mathcal{L}_{\text{CTC}}$) or Cross-Entropy ($\mathcal{L}_{\text{CE}}$) loss, depending on the model architecture. The latter is the alignment loss $\mathcal{L}_{W_2}$ introduced in Equation~\ref{eq:1}.

\paragraph{FADA Guided Decoding Framework (\texttt{FADA}\textsubscript{GD})}

Our proposed method also supports an inference-time application via a guided decoding algorithm that aligns predicted character-level frequency distributions with empirical targets. While training-time alignment enhances robustness to distributional shifts, inference-time adjustment provides an additional mechanism for refining predictions \textit{without} requiring model retraining. To enable this, we incorporate a frequency-aware scoring component into the beam search process, which biases candidate sequences toward those whose character distributions more closely match the expected empirical ones. In this way, we correct potential distributional mismatches at inference time, thereby improving overall transcription accuracy.

In standard beam search, candidate sequences are ranked based on their cumulative log-probability, favoring highly probable continuations. We extend this decoding objective by introducing a penalty term that quantifies the divergence between a candidate sequence’s predicted character frequency distribution and the corresponding empirical distribution. This penalty term is applied at each decoding step, enabling the beam search process to dynamically incorporate character-level alignment as the sequence is generated.
Specifically, at decoding step $t$, each candidate sequence $V = (v_1, v_2, \dots, v_t)$ is scored using the following frequency-aware objective:

\begin{align*}
    \mathcal{S}_{\text{GD}} = & \ [ \lambda \cdot \sum_{i=1}^{t} \log P(v_i | v_{< i}) - \\ 
    & (1 - \lambda) \cdot W_2 (D_{\text{pred}}, D_{\text{emp}})]
\label{eq:2}
\tag{2}
\end{align*}

Here, $P(v_i | v_{< i})$ denotes the model-assigned probability of character $v_i$  given the preceding sequence $v_{<i}$. $D_\text{pred}$ and $D_\text{emp}$ represent the predicted and empirical character frequency distributions, respectively. The first term captures the standard beam search objective and the second imposes a penalty based on the Wasserstein distance $W_2$ (see Equation~\ref{eq:1}). The hyperparameter $\lambda \in [0, 1]$ controls the trade-off between maximizing likelihood and enforcing distributional alignment.

This formulation ensures that candidate sequences with both high likelihood and strong agreement with the expected character frequency distribution are favored during decoding. Importantly, at each decoding step, the alignment penalty is computed over the entire (potentially incomplete) candidate sequence, rather than applied at the individual token level. While a per-token penalty may seem intuitive, it treats frequency alignment as a local property and risks biasing the model toward optimizing each next-token prediction in isolation. In contrast, our approach promotes global coherence by encouraging sequences whose overall character-level statistics align better with the empirical target distribution.  As a result, the guided decoding strategy functions as a lightweight calibration mechanism that leverages known statistical properties of the target domain to enhance recognition performance without additional training. Algorithm~\ref{alg:fada_tr} presents both our training-time and our inference-time alignment framework in pseudo-code form.

\begin{algorithm}
\caption{The proposed FADA framework}\label{alg:fada_tr}
\begin{algorithmic}
\State \LeftComment{2}{Train set w/ domain labels $d_i$ (e.g., century):}
\State $\mathcal{D} = \{(x_i, y_i, d_i)\}$

\vspace{0.8em}

\State \LeftComment{2}{Target character distribution per domain $d$:}
\State $\{p^{(d)}_{\text{target}}\}$ 

\vspace{0.8em}

\State \begin{tabular}{@{}ll@{}}
$\triangleright$ Model with parameters $\theta$: & $f_\theta$
\end{tabular}

\vspace{0.8em}

\State \begin{tabular}{@{}ll@{}}
$\triangleright$ Loss weighting factor: & $\lambda$
\end{tabular}

\vspace{0.8em}

\State \LeftComment{2}{Character Distribution calculation function:}
\State $\text{CharDist}$ 

\Algphase{\small Frequency-Aware Training (\texttt{FADA}\textsubscript{TR})}
\For{each minibatch $\mathcal{B} = \{(x_i, y_i, d_i)\} \subset \mathcal{D}$}
    \State $y_{\text{pred}} \gets f_\theta(x)$ \Comment{Model predictions}

    \State $\mathcal{L}_{\text{CE}} \gets \text{CE}(y_{\text{pred}}, y)$ \Comment{Cross-entropy loss}

    \vspace{0.5em}

    \State \LeftComment{2}{Compute alignment loss across samples}
    \State $\mathcal{L}_{\text{FADA}} \gets 0$
    \For{each $(y_i^{\text{pred}}, d_i)$ in $(y_{\text{pred}}, d)$}
        \State $q_i^{\text{pred}} \gets \text{CharDist}(y_i^{\text{pred}})$
        \State $\mathcal{L}_{\text{FADA}} \mathrel{+}= W_2(q_i^{\text{pred}}, p^{(d_i)}_{\text{target}})$
    \EndFor
    \State $\mathcal{L}_{\text{FADA}} \gets \mathcal{L}_{\text{FADA}} / |\mathcal{B}|$

    \vspace{0.5em}

    \State \LeftComment{2}{Combine losses}
    \State $\mathcal{L}_{\text{total}} \gets \lambda \cdot \mathcal{L}_{\text{CE}} + (1 - \lambda) \cdot \mathcal{L}_{\text{FADA}}$

    \State $\theta \gets \theta - \eta \cdot \nabla_\theta \mathcal{L}_{\text{total}}$ \Comment{Gradient update}
\EndFor

\vspace{1em}
\Algphase{\small Frequency-Aware Guided Decoding (\texttt{FADA}\textsubscript{GD})}
\For{each beam search decoding step}
    \State beamScores = dict()
    \For{each beam search sequence $h$}
        \State $d \gets d_i$ \Comment{\text{e.g., target century/region}}
        
        \vspace{0.6em}
        
        \State \LeftComment{2}{Get LM decoder’s score for h}
        \vspace{0.2em}

        \State $\mathcal{S}_{\text{D}}(h) = \sum_{i=1}^{|h|} \log P(h_i | h_{< i})$

        \vspace{0.6em}

        \State \LeftComment{2}{Compute alignment score (Eq.~\ref{eq:2})} 
        \State $q_h \gets \text{CharDist}(h)$
        \State $\mathcal{S}_{\text{align}}(h) \gets W_2(q_h, p^{(d)}_{\text{target}})$

        \vspace{0.6em}

        \State \LeftComment{2}{Compute \texttt{FADA}\textsubscript{GD} score}
        \State $\mathcal{S}_{\text{GD}}(h) \gets \lambda \cdot \mathcal{S}_{\text{D}}(h) - (1-\lambda) \cdot \mathcal{S}_{\text{align}}(h)$

        \vspace{0.6em}

        \State \LeftComment{2}{Update beam score}
        \State $\text{beamScores}[h] \gets \text{S}_{\text{GD}}(h)$
    \EndFor
\EndFor
\end{algorithmic}
\end{algorithm}

\section{Experiments}
\label{sec:experiments}
We evaluate the performance of \texttt{FADA} primarily on HTR, using two model architectures across three datasets, two real-world and one synthetic, spanning Greek and French manuscripts from the 10\textsuperscript{th} to 16\textsuperscript{th} cent. CE. To demonstrate the broader applicability of our method, we also conduct secondary experiments on Automatic Speech Recognition (ASR) using the Whisper model \cite{whisper} and an English dataset. Next, we discuss dataset and model details (\S\ref{subsec:datasets}-\S\ref{subsec:models}), baselines (\S\ref{subsec:baselines}), evaluation metrics (\S\ref{subsec:eval_measures}), and then we present the experimental results (\S\ref{subsec:results}), followed by an ablation study and qualitative analyses.

\subsection{Datasets}
\label{subsec:datasets}

\noindent \textbf{\texttt{HPGT} Dataset} (HTR): \texttt{HPGT} \cite{platanou-etal-2022-handwritten} is a Handwritten Paleographic Greek Text recognition dataset comprising 77 digitized page-level images from Greek manuscripts in the Oxford University Bodleian Library.\footnote{\scriptsize\url{https://bav.bodleian.ox.ac.uk/greek-manuscripts}} The images are segmented into 1,698 line-level between the 10\textsuperscript{th} to 16\textsuperscript{th} century CE. See Appendix~\ref{sec:appendix-hpgt} for more details about the intra-dataset character distributions and correlations. 

\noindent \textbf{\texttt{CATMuS-FR} Dataset} (HTR): \texttt{CATMuS} \cite{catmus} is a multilingual manuscript dataset containing over 200 historical texts. We extracted all French-written samples to form \texttt{CATMuS-FR}, consisting of 3,401 samples spanning four centuries (13\textsuperscript{th} to 16\textsuperscript{th}) and six distinct script types. See Appendix~\ref{sec:appendix-catmus} for more details about the intra-dataset character distributions and correlations. 

\noindent \textbf{Synthetic Dataset} (HTR): To evaluate \texttt{FADA} under controlled conditions, we generated a synthetic Greek dataset with deliberately induced shifts in character frequency distributions. Using predefined prompts and topics, we synthesized text with four different LLMs (GPT-4 \cite{gpt4}, Llama-3.1 \cite{grattafiori2024llama3herdmodels}, Gemini \cite{gemini}, and Claude-3.5 \cite{claude2024}) each producing outputs with distinct statistical properties, effectively simulating four separate ``centuries.'' A total of 1,690 text segments were carefully curated and rendered as images using an open-source handwritten text generator.\footnote{\scriptsize\url{https://github.com/Belval/TextRecognitionDataGenerator}} More details about the dataset creation are provided in Appendix~\ref{sec:appendix-synth}.

\noindent \textbf{\texttt{EdAcc} Dataset} (ASR): \texttt{EdAcc} \cite{sanabria23edacc} is an ASR dataset containing 40 hours of transcribed speech from speakers with diverse linguistic backgrounds and English accents. We selected the four most represented categories; US English, Southern British English, Irish English, and Indian English, resulting in 2,756 samples.
More details about the inter-group character distribution shifts and correlation scores are provided in Appendix~\ref{sec:appendix-edacc}.

\subsection{Backbone Models}
\label{subsec:models}

We experiment with three backbone models featuring distinct architectures across two recognition tasks. In our experiments, C-RNN is trained from scratch, while TrOCR and Whisper are fine-tuned from publicly available pretrained checkpoints. We evaluate each model under two baseline configurations (see \S\ref{subsec:baselines}) as well as our proposed distributional alignment framework (\texttt{FADA}).
 
\noindent \textbf{C-RNN} (HTR): C-RNN \cite{crnn} is a widely used architecture for HTR that combines a CNN for visual feature extraction with an RNN for sequential text generation. The model outputs text at character level and is trained using the CTC loss.

\noindent \textbf{TrOCR} (HTR): TrOCR \cite{trocr} is a state-of-the-art Transformer-based \cite{Vaswani2017AttentionIA} approach for HTR. It comprises a Vision Transformer (ViT) \cite{vit} as an image encoder and a text Transformer \cite{Vaswani2017AttentionIA} as the decoder. TrOCR operates at token level and is trained using CE loss.

\noindent \textbf{Whisper} (ASR): Whisper \cite{whisper} is a state-of-the-art speech recognition model based on a Transformer-based encoder-decoder architecture. It is pre-trained on a range of speech processing tasks, including ASR and speech translation.

\subsection{Baseline Configurations}
\label{subsec:baselines}

\smallskip
\textbf{Fine-tuned Backbone Model:}
This configuration serves as our simplest baseline. For Transformer-based models (TrOCR and Whisper), we fine-tune publicly available pre-trained checkpoints on each target dataset using only the standard task-specific objective (CE loss). For the C-RNN model, which does not rely on a pre-trained backbone, we train the model from scratch using CTC loss. This setup reflects the default recognition setting without any frequency-level supervision, and enables direct assessment of the added value introduced by our proposed distributional alignment mechanism.

\smallskip
\noindent\textbf{Temporal/Regional Token Tagging (TR-TT)}: Following strategies employed in multilingual and multi-task models such as mBART \cite{liu2020multilingual} and Whisper \cite{whisper}, we fine-tune each backbone model by prepending a special token to each input sequence indicating the temporal or regional context. This token is treated as part of the input and provides an implicit conditioning signal for generation. This configuration does not incorporate any explicit alignment with empirical distributions. Instead, it relies on the model's ability to learn contextual associations from the data. Training is performed end-to-end using only the standard task-specific objective (CTC or CE).

\begin{table*}[t]
\centering
\small
\setlength{\tabcolsep}{8pt}
\begin{tabular}{llcccccc}
\toprule
\multirow{2}{*}{\textbf{Dataset}} & \multirow{2}{*}{\textbf{Model}} & \multicolumn{2}{c}{\textbf{Standard FT}} & \multicolumn{2}{c}{\textbf{FT w/ TR-TT}} & \multicolumn{2}{c}{\textbf{FT w/ FADA\textsubscript{TR-GD} (Ours)}} \\
\cmidrule(lr){3-4} \cmidrule(lr){5-6} \cmidrule(lr){7-8}
 & & CER ↓ & WER ↓ & CER ↓ & WER ↓ & CER ↓ & WER ↓ \\
\midrule

\multirow{2}{*}{HPGT} 
  & TrOCR & 26.92 \tiny{(0.96)} & 73.82 \tiny{(1.31)} & 26.89 \tiny{(0.61)} & 72.53 \tiny{(0.81)} & \textbf{25.06} \tiny{(0.91)} & \underline{70.14} \tiny{(0.48)} \\
  & C-RNN & 24.72 \tiny{(0.38)} & 77.08 \tiny{(0.85)} & 24.41 \tiny{(0.55)} & 76.66 \tiny{(0.81)} & \textbf{23.89} \tiny{(0.34)} & \underline{76.31} \tiny{(0.77)} \\
\midrule

\multirow{2}{*}{CATMuS} 
  & TrOCR & \textbf{9.56} \tiny{(0.22)} & 37.02 \tiny{(0.28)} & 9.68 \tiny{(0.17)} & \underline{37.01} \tiny{(0.69)} & {9.63} \tiny{(0.25)} & 37.05 \tiny{(0.47)} \\
  & C-RNN & 17.08 \tiny{(0.29)} & 54.79 \tiny{(0.32)} & 27.12 \tiny{(1.13)} & 67.61 \tiny{(1.98)} & \textbf{16.79} \tiny{(0.52)} & \underline{53.65} \tiny{(1.02)} \\
\midrule

\multirow{2}{*}{Synthetic} 
  & TrOCR & 5.94 \tiny{(0.09)} & 17.61 \tiny{(0.31)} & 5.48 \tiny{(0.17)} & 17.12 \tiny{(1.96)} & \textbf{5.19} \tiny{(0.27)} & \underline{16.60} \tiny{(0.45)} \\
  & C-RNN & 9.71 \tiny{(0.73)} & 31.58 \tiny{(2.73)} & 9.82 \tiny{(0.34)} & 34.58 \tiny{(1.78)} & \textbf{8.31} \tiny{(0.12)} & \underline{27.08} \tiny{(0.36)} \\

\bottomrule
\end{tabular}
\caption{
\textbf{Main evaluation results.} We report average CER and WER (across three runs) and the SEM (in parentheses next to each score) across all datasets and models. FT indicates fine-tuning the model. Our proposed method outperforms all baselines in most cases.  FADA\textsubscript{TR-GD} denotes our proposed approach combining training-time alignment and inference-time guided decoding. TR-TT is a token-based contextual conditioning baseline described in \S\ref{subsec:baselines}. The best CER and WER scores per model and dataset are in bold and underlined respectively.
}
\label{tab:main_results}
\end{table*}

\subsection{Evaluation Metrics}
\label{subsec:eval_measures}


\textbf{Character Error Rate (CER)} measures transcription accuracy at character level. It is computed as the edit distance between the predicted and the reference text, normalized by the length of the reference. Lower CER indicates better performance.

\noindent\textbf{Word Error Rate (WER)} is the word-level counterpart of CER. It calculates the edit distance between the predicted and reference word sequences, normalized by the number of words in the reference. As with CER, lower WER values correspond to higher transcription accuracy.

\subsection{Experimental Results}
\label{subsec:results}

We evaluate our proposed method (\texttt{FADA}) across multiple backbone models (\S\ref{subsec:models}) and compare it to two baseline configurations, each involving fine-tuning (or training from scratch depending on the architecture) the backbone models on the target dataset (\S\ref{subsec:baselines}). Our main configuration, dubbed \texttt{FADA}\textsubscript{TR-GD}, fine-tunes (or trains) each model using a combined objective that includes the standard task-specific loss (CTC or CE) and our distributional alignment loss. At inference time, we apply our guided decoding algorithm to further refine outputs. This setup integrates frequency alignment at both training and inference stages, resulting in \textit{end-to-end frequency-aware} recognition. We run each experiment three times with different seeds and report the mean and standard error of the mean (SEM) for each metric (see \S\ref{sec:discussion} for reproducibility).

\begin{table}[t]
\small
\centering
\setlength{\tabcolsep}{6pt}
\renewcommand{\arraystretch}{1.1}
\begin{tabular}{lccc}
\toprule
\textbf{Model} & \textbf{Standard FT} & \textbf{w/ TR-TT} & \textbf{w/ FADA\textsubscript{TR-GD}} \\
\midrule
Whisper & 27.11 / 40.02 & 28.88 / 44.09 & \textbf{26.70} / \underline{38.02} \\
\bottomrule
\end{tabular}
\caption{\textbf{ASR results on EdAcc.} CER / WER for Whisper-small across three configurations. Best scores per metric are bolded and underlined respectively.}
\label{tab:asr_results}
\end{table}

\paragraph{Quantitative Analysis} Table~\ref{tab:main_results} reports the performance of the two models on the three HTR datasets. For each model (TrOCR, C-RNN), we report three scores per evaluation metric (CER, WER); one for each baseline approach, and one for our proposed method ($\texttt{FADA}_\text{TR-GD}$). For the latter, scores obtained with the best hyperparameter $\lambda$ are reported. In Table~\ref{tab:asr_results}, we examine the effectiveness of our proposed method in ASR. We report the performance of Whisper on the Edacc dataset across the two baseline methods and $\texttt{FADA}_\text{TR-GD}$.

The results in both tables show that \texttt{FADA}\textsubscript{TR-GD} consistently improves both text and speech recognition performance across a range of datasets and model architectures, demonstrating the effectiveness of character-level frequency distribution alignment in recognition tasks. Despite the single case where \texttt{FADA}\textsubscript{TR-GD} does not improve over TrOCR baseline (falling within the SEM) it generally provides substantial gains (up to three WER units), highlighting its robustness and broad applicability. These findings confirm that aligning model predictions with empirical character statistics can enhance recognition accuracy, even under diverse temporal or regional shifts.

\begin{table*}[h]
    \centering
    \small
    \renewcommand{\arraystretch}{1.2}
    \begin{tabular}{|p{2.4cm}||p{8.0cm}|p{1.3cm}|p{1.3cm}|}
        \hline
        \centering\textbf{Method} & \centering\textbf{Transcription} & \centering\textbf{CER} & \hspace{0.18cm}\textbf{WER} \\ 
        \hline
        \centering\small Baseline (TrOCR) & \small\textgreek{\textcolor{darkgreen}{φων}η\textcolor{darkgreen}{ς} \textcolor{darkgreen}{εἰσπ}νατιον τὴν δὲ κας η\textcolor{darkgreen}{ν} \textcolor{darkgreen}{συμφ}όνας \textcolor{darkgreen}{ἀφ}η\textcolor{darkgreen}{καν} \textcolor{darkgreen}{τήν}} & \centering29.16 & \small\hspace{0.18cm} 1.14 \\ 
        \centering\small \texttt{FADA}\textsubscript{GD} & \small\textgreek{\textcolor{darkgreen}{φων}η\textcolor{darkgreen}{ς} \textcolor{darkgreen}{εἰσπ}νατιον τὴν δὲ κας η\textcolor{darkgreen}{ν} \textcolor{darkgreen}{συμφ}όνας \textcolor{darkgreen}{ἀφ}η\textcolor{darkgreen}{καν} \textcolor{darkgreen}{τήν}} & \centering29.16 & \small\hspace{0.18cm} 1.14 \\ 
        \centering\small \texttt{FADA}\textsubscript{TR} & \small\textgreek{\textcolor{darkgreen}{φων}η\textcolor{darkgreen}{ς} \textcolor{darkgreen}{εἰσπ}να\textcolor{darkgreen}{ττονται δίκας} ἡ\textcolor{darkgreen}{ν συμφώνως ἀφ}η καντὴν} & \centering\textbf{14.58} & \small\hspace{0.18cm} \textbf{0.71} \\ 
        \centering\small \texttt{FADA}\textsubscript{TR-GD} & \small\textgreek{\textcolor{darkgreen}{φων}η\textcolor{darkgreen}{ς} \textcolor{darkgreen}{εἰσπ}να\textcolor{darkgreen}{ττονται δίκας} ἡ\textcolor{darkgreen}{ν συμφώνως ἀφ}η καντὴν} & \centering\textbf{14.58} & \small\hspace{0.18cm} \textbf{0.71} \\
        \hline
        \centering\small \textit{Ground Truth} & \small{\textgreek{φωνῆς εἰσπράττονται δίκας ἢν συμφώνως ἀφῆκαν τήν}} & & \\ 
        \hline
    \end{tabular}
    \\
    \begin{tabular}{|p{2.4cm}||p{8.0cm}|p{1.3cm}|p{1.3cm}|}
        \hline
        \centering\small Baseline (Whisper) & \small\textcolor{darkgreen}{yeah} and all \textcolor{darkgreen}{i bet} and \textcolor{darkgreen}{like in the apartment that we have} & \centering18.18 & \small\hspace{0.18cm} 0.23 \\ 
        \centering\small \texttt{FADA}\textsubscript{GD} & and uh \textcolor{darkgreen}{i bet} in \textcolor{darkgreen}{like in the apartment that we have} & \centering19.98 & \small\hspace{0.18cm} 0.30\\ 
        \centering\small \texttt{FADA}\textsubscript{TR} & \small and uh \textcolor{darkgreen}{i bet} in \textcolor{darkgreen}{like in the apartment that we have} & \centering19.98 & \small\hspace{0.18cm} 0.30\\ 
        \centering\small \texttt{FADA}\textsubscript{TR-GD} & \small\textcolor{darkgreen}{yeah i know i bet} in \textcolor{darkgreen}{like in the apartment that we have} & \centering\small\textbf{3.63} & \hspace{0.26cm}\small\textbf{0.07} \\
        \hline
        \centering\small \textit{Ground Truth} & \small {yeah i know i bet uh like in the apartment that we have} & & \\ 
        \hline
    \end{tabular}
    \caption{Qualitative analysis of \texttt{FADA} on HTR and ASR tasks. The upper section presents a Greek manuscript transcription example, while the lower section showcases an English ASR transcript. Correctly restored characters are highlighted in green. CER and WER values are reported for each setting, demonstrating the improvements.}
    \label{tab:qualitative}
\end{table*}

\begin{table}[t]
\centering
\small
\setlength{\tabcolsep}{4.5pt}
\begin{tabular}{llccc}
\toprule
\textbf{Dataset} & \textbf{Model} & \textbf{FADA\textsubscript{GD}} & \textbf{FADA\textsubscript{TR}} & \textbf{FADA\textsubscript{TR-GD}} \\
\midrule
\multirow{2}{*}{HPGT} 
  & TrOCR  & $-$1.4\% & $-$6.5\% & \textbf{$-$6.9\%} \\
  & C-RNN  & $-$0.8\% & $-$3.2\% & \textbf{$-$3.3\%} \\
\midrule
\multirow{2}{*}{CATMuS} 
  & TrOCR  & $+$1.7\% & $+$1.6\% & \textbf{$+$0.7\%} \\
  & C-RNN  & $-$0.4\% & $-$1.1\% & \textbf{$-$1.7\%} \\
\midrule
\multirow{2}{*}{Synthetic} 
  & TrOCR  & $-$0.2\% & $-$10.6\% & \textbf{$-$12.6\%} \\
  & C-RNN  & $-$0.1\% & $-$12.9\% & \textbf{$-$14.4\%} \\
\bottomrule
\end{tabular}
\caption{
\textbf{Ablation Study: The relative improvement in CER over standard fine-tuning across \texttt{FADA} configurations.} Negative values indicate error reduction. \textbf{Bold} marks the best per model/dataset (lower is better).
}
\label{tab:fada_ablation}
\end{table}

\smallskip\noindent\textbf{Ablation study}
In addition to our full end-to-end configuration (\texttt{FADA}\textsubscript{TR-GD}), incorporating alignment at both training and inference, we assess two partial variants: training-only alignment (\texttt{FADA}\textsubscript{TR}) and inference-only alignment via guided decoding (\texttt{FADA}\textsubscript{GD}). As can be seen in Table~\ref{tab:fada_ablation}, inference-time alignment improves recognition performance in most cases, despite operating solely as a post-hoc adjustment to model outputs (see also \S\ref{sec:discussion}). Training-time alignment yields larger improvements overall, as it directly shapes the model’s internal representations by reducing character frequency discrepancies during optimization. While the combined approach (\texttt{FADA}\textsubscript{TR-GD}) consistently achieves the best overall performance, training-only and inference-only alignment offer meaningful improvements at a lower computational cost.

\smallskip\noindent\textbf{Qualitative Analysis} Table~\ref{tab:qualitative} presents two transcription examples, one from each task, to illustrate the qualitative impact of our proposed framework. We compare outputs from the fine-tuned backbone model (TrOCR or Whisper) with those produced by the same model enhanced with inference-time alignment (\texttt{FADA}\textsubscript{GD}), training-time alignment (\texttt{FADA}\textsubscript{TR}), and the full configuration combining both (\texttt{FADA}\textsubscript{TR-GD}). 
In both examples, the full configuration (\texttt{FADA}\textsubscript{TR-GD}) produced the most accurate transcriptions, closely matching the ground truth and outperforming the standard fine-tuned model as well as the individual \texttt{FADA} variants. 

In the first case (HPGTR dataset, HTR), training-time alignment (\texttt{FADA}\textsubscript{TR}) alone achieved the same CER and WER as the full configuration, suggesting that it can often be sufficient. In contrast, the second example (EdAcc dataset, ASR) illustrates a complementary effect; while neither training-only nor inference-only improved upon the baseline in isolation, their combination resulted in a significantly better transcription. This result highlights the synergistic benefit of combining improved representations (from training-time alignment) with frequency-aware decoding (via inference-time alignment). 
Note that while \texttt{FADA} is not designed to explicitly optimize for WER, improvements in CER which is our primary objective, naturally translate to better WER in most cases. In our examples, this correlation is clearly reflected, further reinforcing the robustness of our alignment-based framework.

\smallskip\noindent\textbf{Per-Century/Region Results} To further assess the effectiveness of \texttt{FADA}, we evaluate model performance in terms of CER across individual centuries (for HTR) and linguistic regions (for ASR). Rather than reporting only aggregated scores, this analysis focuses on how well each model performs on specific temporal and regional subsets. The results reveal substantial variation in difficulty across these subsets. Notably, \texttt{FADA}-enhanced models (\texttt{FADA}\textsubscript{TR}, \texttt{FADA}\textsubscript{GD}, and \texttt{FADA}\textsubscript{TR-GD}) exhibit improved and more consistent performance across different subsets compared to standard fine-tuning, effectively narrowing the gap between easier and more challenging cases. A detailed breakdown of per-century and per-region results is provided in Appendix~\ref{sec:appendix-per-century}.

\noindent\textbf{Lipogram Generation} To test \texttt{FADA}’s applicability beyond text recognition, we apply it to lipogram generation; a constrained generation task where the model must avoid generating a given (forbidden) character \cite{lipograms}. We prompted Llama-3.1 with 30 diverse topics and applied our proposed guided decoding algorithm (\texttt{FADA}\textsubscript{GD}) by setting the target frequency of the forbidden character to zero (i.e., the model must not generate it). We repeated this process for three different forbidden characters and compared standard beam search with \texttt{FADA}\textsubscript{GD} decoding, measuring violation rate and perplexity. Results show that \texttt{FADA} effectively enforces character-level constraints during open-ended generation. Full results, topic list, and prompts appear in Appendix \ref{sec:appendix-lipograms}.

\smallskip\noindent\textbf{Beam size sensitivity study} Undertaking a sensitivity study, we investigate the effect of beam size, comparing the performance of the fine-tuned backbone model with and without training-time alignment (i.e., $\texttt{FADA}_\textsubscript{TR}$). We examine beam sizes of $1$, $5$ and $7$. Increasing the beam size in the standard fine-tuned TrOCR model yields only minor gains. In contrast, $\texttt{FADA}\textsubscript{TR}$ shows stronger improvements, especially at beam size of $1$, where it consistently outperforms the baseline even with a beam size of $7$. This shows that frequency-aware training boosts transcription quality without relying on large beams, with smaller returns as beam size grows. For more results, see Appendix \ref{sec:appendix-beam-ablation}.

\section{Discussion}
\label{sec:discussion}


\paragraph{Importance of inference-time alignment} Table~\ref{tab:fada_ablation} shows that inference-time alignment (\texttt{FADA}\textsubscript{GD}) yields smaller gains compared to training-time alignment (\texttt{FADA}\textsubscript{TR}), which is expected since the latter can update model parameters. However, inference-time alignment offers several practical advantages: it is computationally lightweight, requires no retraining, and is ideal for low-resource scenarios. It also acts as a post-hoc calibration step, often improving results when combined with \texttt{FADA}\textsubscript{TR} (Table~\ref{tab:fada_ablation}), suggesting that frequency-aware scoring can refine outputs even when the underlying model has already been trained to align with frequency statistics. Finally, it supports zero-shot adaptation, as is shown in our Lipogram experiment (\S\ref{sec:experiments}), where it effectively guides generation away from forbidden characters without any fine-tuning.

\paragraph{Advantages over Temporal/Regional Tagging (TR-TT Baseline)} Contextual-conditioning strategies, such as prepending special tokens to indicate conditioning attributes (e.g., language or time) are widely used in recognition tasks \cite{liu2020multilingual}. While effective, these methods control the model in an implicit way that is hard to interpret or adjust. In contrast, by explicitly aligning predicted character distributions with empirical ones, computed from the training data, \texttt{FADA} makes the alignment process interpretable and transparent. Also, it gives control to users over the output, who can modify the target distribution, encouraging (or discouraging) certain characters.

\paragraph{Differences from Domain Adaptation Methods} Traditional domain adaptation methods typically assume that a model is trained on a single source-domain (e.g., modern printed text) and then adapted to a different target-domain with limited labeled data, using techniques such as feature alignment or adversarial training. In contrast, \texttt{FADA} is not designed to transfer between domains, but to handle \emph{intra-dataset distributional variation} within a single dataset. Instead of treating each subset (e.g., century, region) as an isolated domain, we train a single model on the entire dataset and guide it to respect the empirical character distribution associated with each training sample. This allows the model to learn general task-relevant representations while adjusting its output to reflect expected patterns of the specific input context without requiring separate adaptation stages or domain boundaries.

\paragraph{Practicality and Feasibility of Domain-specific Character Frequency Priors} Both our training-time alignment and inference-time guided decoding methods rely on access to domain-specific intra-dataset character frequency distributions. While such information may not always be available, our focus is on settings where these distributions are relatively stable and can be approximated from external sources. For example, in historical text or regional speech datasets, the set of frequently used characters tends to shift gradually over time or geography, making it feasible to estimate frequency distributions using linguistic studies or domain-specific corpora. We therefore propose FADA as a flexible framework, where alignment can be applied when such priors are available or reasonably inferred, rather than assuming strict access to domain labels at all times.

\paragraph{Applicability to Higher-order Linguistic Patterns} Although FADA is formulated as a unigram-level alignment method, it is inherently extensible to higher-order linguistic patterns (e.g. bigrams). One approach involves directly matching predicted n-gram frequency distributions to known domain-specific n-gram priors, encouraging the model to reflect more structured co-occurrence statistics. Alternatively, as we already exploit in our experiments with Transformer-based models, the model may generate subword units, while FADA continues to operate on the decoded character sequence. 

\paragraph{Computational Overhead and Reproducibility} Training with $\texttt{FADA}_\text{TR}$ introduces a modest $4$–$6$\% increase in training time due to the alignment loss computation, while $\texttt{FADA}_\text{GD}$ slows decoding by $7$–$9$\% due to the added frequency-based scoring. The memory overhead is minimal. Given the accuracy improvements and character-level control, this trade-off remains favorable in most cases. All models were trained with fixed seeds to ensure reproducibility. For CTC-based models, we used the deterministic Baidu CTC implementation. Training/fine-tuning ran for up to $30$ epochs with early stopping (patience $= 4$), using a batch size of $4$ and a beam size of $5$. Details on the $\lambda$ hyperparameter tuning are included in Appendix~\ref{sec:appendix-tuning}.


\section{Conclusion}
\label{sec:conclusion}

We present \texttt{FADA} (Frequency-Aware Distribution Alignment), a framework mitigating character relative frequency distribution shifts in text and speech recognition, consisting of two components: (i) a training-time alignment loss minimizing discrepancies between predicted and empirical character distributions; (ii) an inference-time guided decoding frequency-based algorithm that dynamically adjusts predictions. Our experiments on HTR and ASR demonstrate consistent performance improvements across datasets, languages, writing systems, and model architectures.
Directions of future work include extending \texttt{FADA} by incorporating bi-grams and tri-grams, enabling alignment at the sequence level rather than for individual characters (unigrams). Also, we plan to explore case-sensitive frequency distributions, and to improve recognition of named entities and specialized terminology. Finally, we aim to develop adaptive weighting strategies for $\lambda$, allowing a dynamic adjustment instead of relying on a fixed hyperparameter.

\section*{Limitations}
\label{sec:limitations}
While \texttt{FADA} consistently enhances recognition performance across multiple datasets and models, its effectiveness relies on the availability of empirical character frequency distributions. In scenarios where such distributions are not well-defined or highly variable, its impact may be less pronounced. Additionally, the current approach focuses on unigram frequency alignment, leaving room for future exploration of higher-order character dependencies. Although \texttt{FADA} introduces only a modest computational overhead, optimizing its efficiency for real-time applications remains an area of interest. These considerations highlight directions for further refinement rather than fundamental constraints, as \texttt{FADA} remains broadly applicable across different recognition settings.

\section*{Acknowledgments}
This work has been partially supported by project MIS 5154714 of the National Recovery and Resilience Plan Greece 2.0 funded by the European Union under the NextGenerationEU Program.

\bibliography{acl_latex}

\appendix


\section{HPGT Dataset - Analysis}
\label{sec:appendix-hpgt}

\begin{figure}[th!]
    \centering
    \includegraphics[width=.4\textwidth]{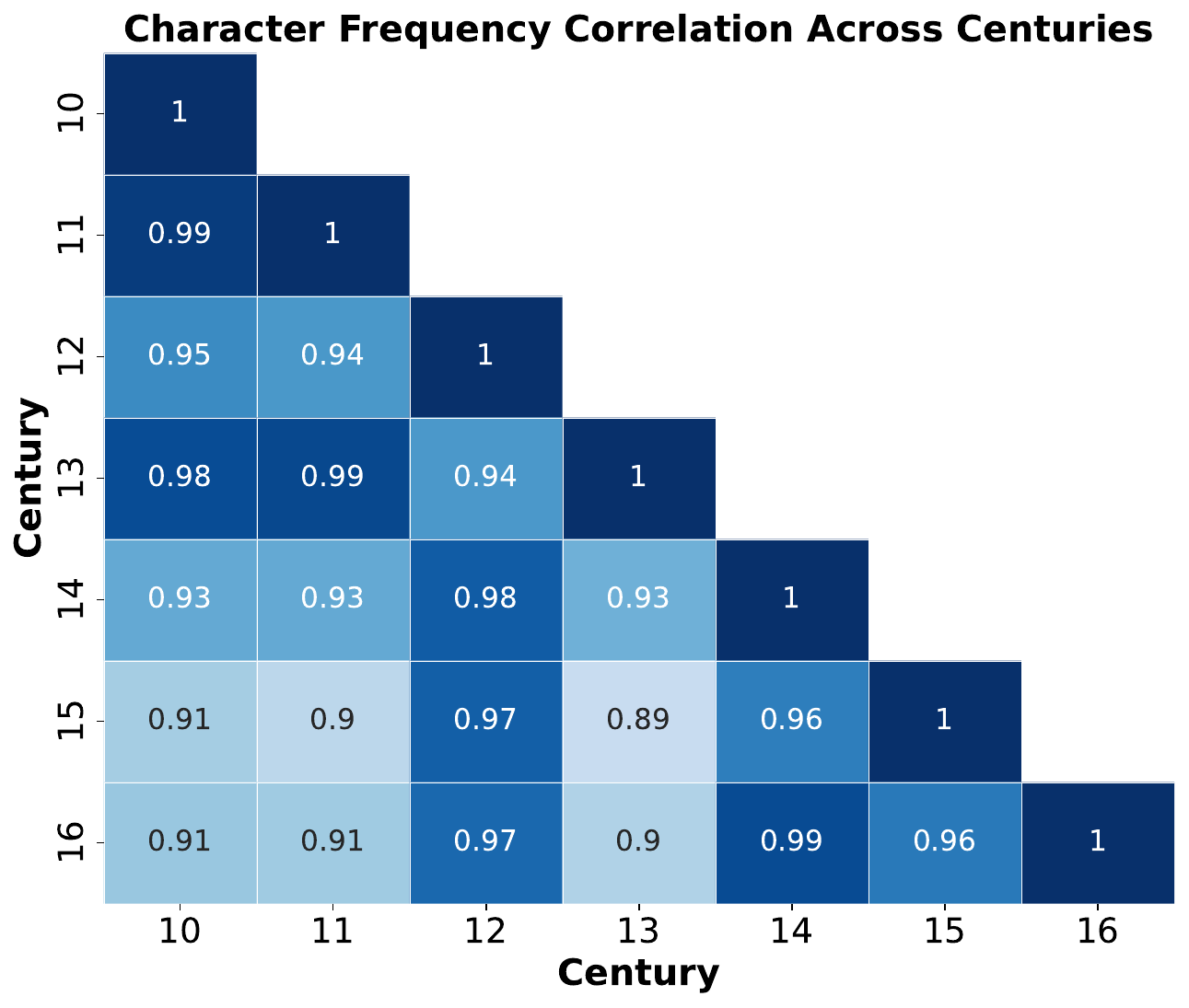}
\caption{Heatmap of Pearson correlation measuring distributional similarity across centuries; darker shades indicate higher correlation.}
\label{fig:2} 
\end{figure}

This section presents a further analysis on the intra-dataset correlation throughout centuries in the HPGT dataset. In Figure~\ref{fig:2}, we quantify the correlation trends between character frequency distributions across centuries, computed using Pearson's correlation ($r$). Adjacent centuries tend to exhibit strong correlations (e.g., 10th–11th: 0.99; 12th–13th: 0.94), whereas distant pairs diverge more substantially (e.g., 10th–16th: 0.91; 13th–15th: 0.89). While correlations above 0.9 generally indicate high similarity, such differences are meaningful in the context of a shared alphabet and may reflect shifts in writing practices, document selection, or biases introduced during data curation. These findings highlight the presence of intra-dataset distribution shifts, whether linguistic or dataset-induced, that can degrade model performance when training ignores subset-specific statistical variation.

\section{CATMuS Dataset - Analysis}
\label{sec:appendix-catmus}

\begin{figure}[th!]
    \centering
    \includegraphics[width=7.8cm, height=8cm, trim=5 5 6 7, clip]{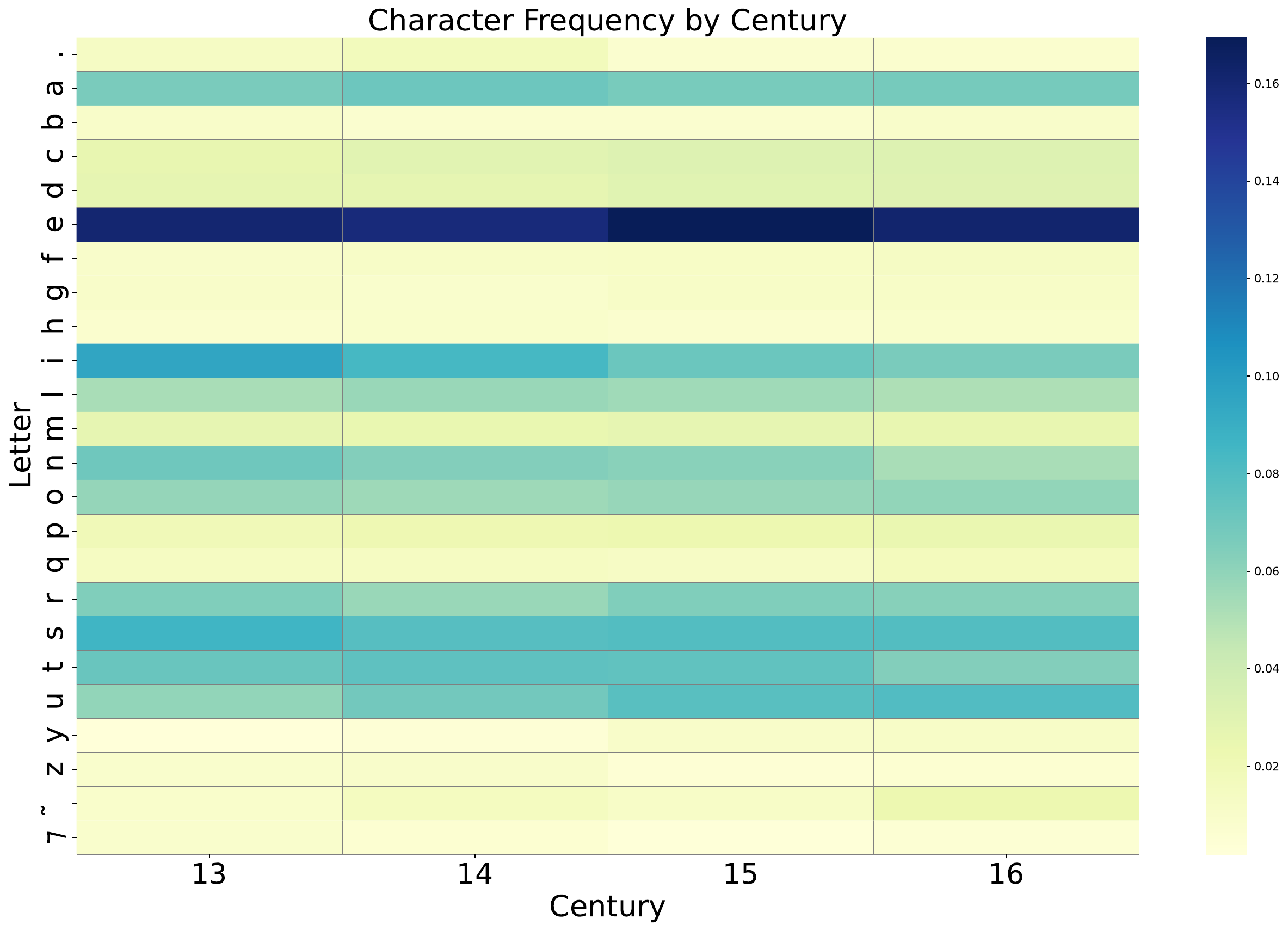}
\caption{Heatmap illustrating the relative frequency distribution of characters in CATMuS \cite{catmus}.}
\label{fig:3} 
\end{figure}

\begin{figure}[th!]
    \centering
    \includegraphics[width=7.5cm, height=10cm, trim=5 5 6 7, clip]{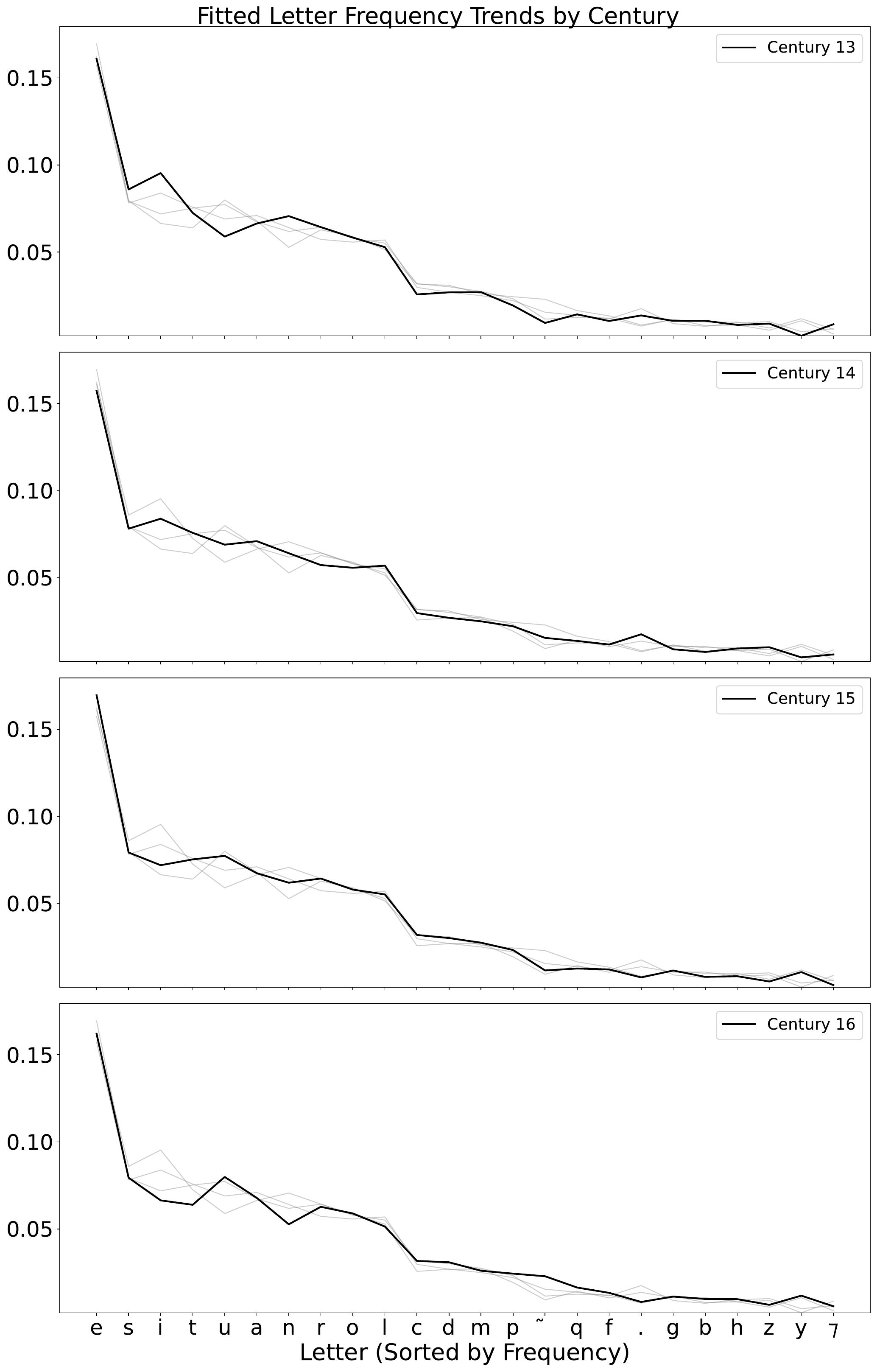}
\caption{Fitted character frequency trends across different centuries in CATMuS \cite{catmus}. Each subplot represents a specific century, with the black line indicating the LOWESS-smoothed letter frequency distribution for that century. Gray lines represent the character frequency trends of the rest of the centuries for comparison.}
\label{fig:4} 
\end{figure}

\begin{figure}[th!]
    \centering
    \includegraphics[width=8cm, height=7cm, trim=6 7 4 26, clip]{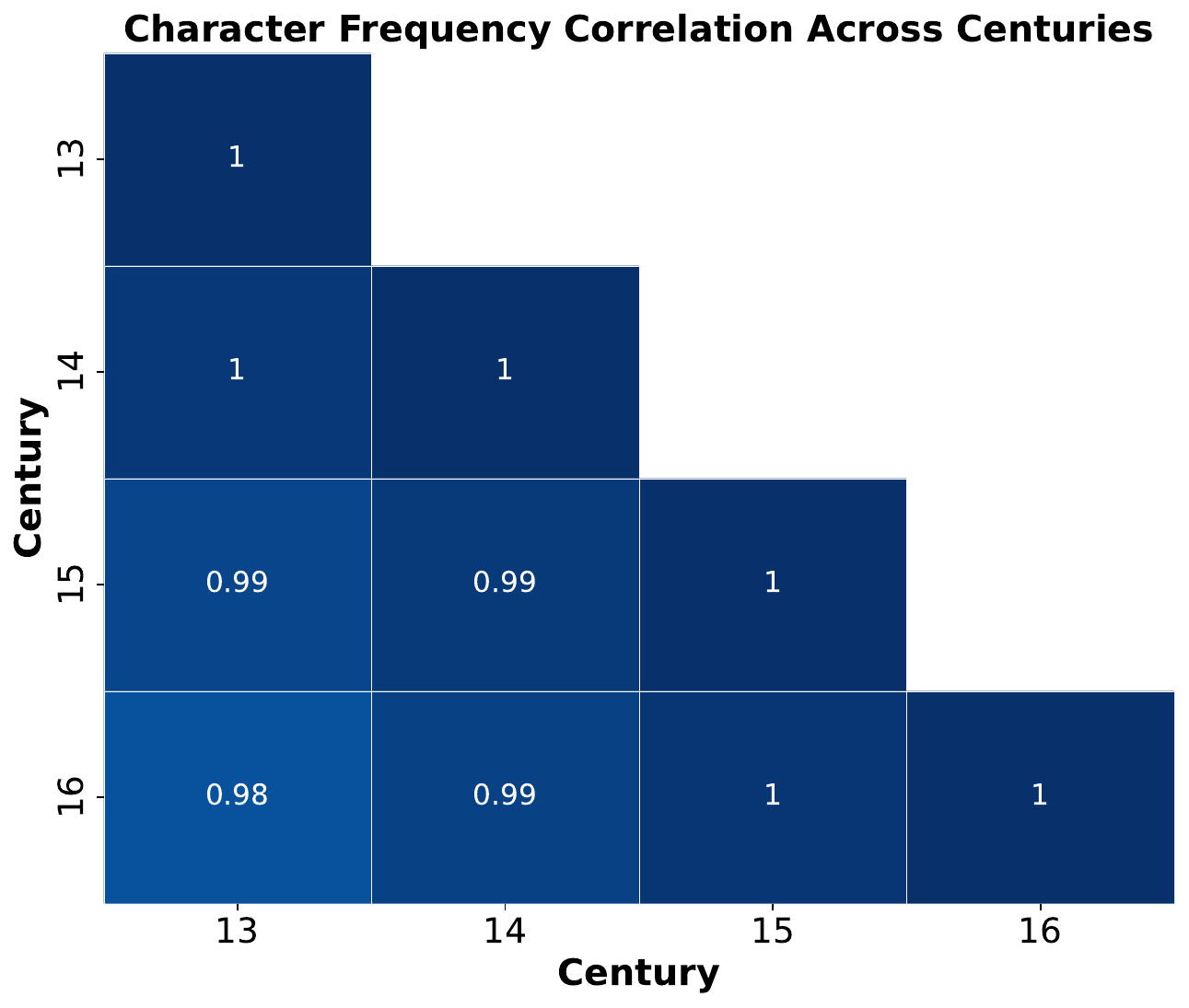}
\caption{Pearson correlation heatmap of the character frequency distribution across centuries in CATMuS \cite{catmus}. Higher values (darker shades) indicate stronger similarity in character usage, while lower values (lighter shades) suggest greater divergence.}
\label{fig:5} 
\end{figure}

In this section, we examine character frequency distribution shifts in the CATMuS dataset \cite{catmus}, complementing the analysis presented for other datasets. Figure~\ref{fig:3} visualizes the relative frequency distributions across four centuries, revealing more subtle intra-dataset variations compared to the rest of the HTR datasets. While some characters exhibit relatively stable frequencies, others show minor fluctuations across different time periods. Figure~\ref{fig:4} further illustrates these trends through LOWESS-smoothed frequency distributions, where the black line represents the character distribution for a given century, and the gray lines provide comparisons to the other centuries. The trends indicate that while there are observable shifts, they are not as pronounced as in datasets covering a broader temporal range. Finally, Figure~\ref{fig:5} presents a Pearson correlation heatmap, quantifying the similarity between character distributions. The consistently high correlation values (e.g., $15$th–$16$th: $0.99$) suggest that distributional shifts are relatively minor, confirming that character frequencies remain largely stable over time in this dataset.

\section{Synthetic Dataset - Analysis}
\label{sec:appendix-synth}

\begin{figure}[th!]
    \centering
    \includegraphics[width=7.8cm, height=8cm, trim=5 5 6 7, clip]{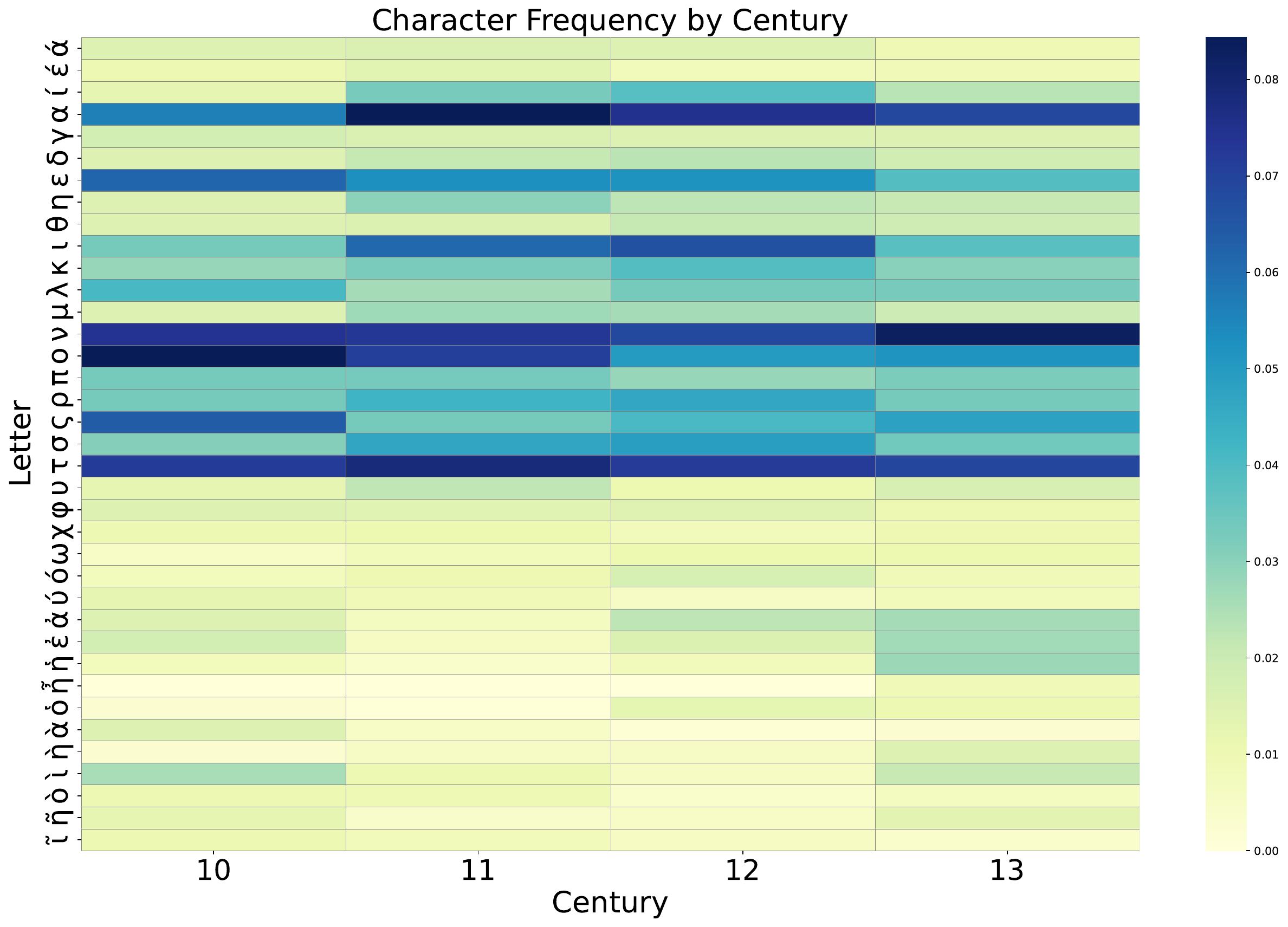}
\caption{Heatmap illustrating the relative frequency distribution of Greek characters in our synthetic dataset (see  \S\ref{subsec:datasets}).}
\label{fig:6} 
\end{figure}

Similarly, we provide an analysis for our synthetic dataset, which was designed to exhibit controlled character frequency shifts across centuries. Figure~\ref{fig:6} presents a heatmap of character usage, revealing stable trends for some characters and substantial fluctuations for others.

\begin{figure}[th!]
    \centering
    \includegraphics[width=7.0cm, height=9.5cm, trim=5 5 6 7, clip]{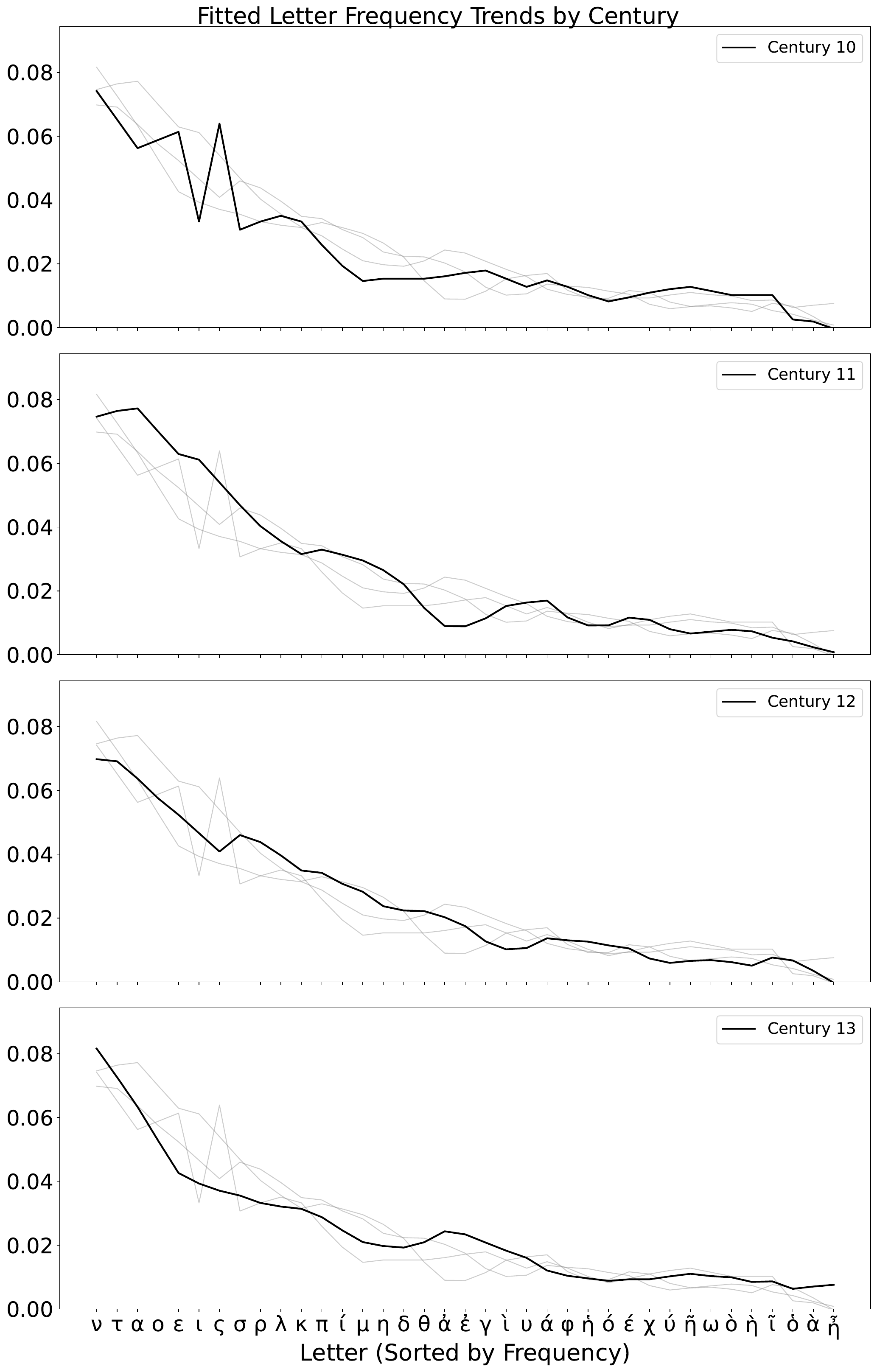}
\caption{Fitted character frequency trends across different centuries. Each subplot represents a specific century, with the black line indicating the LOWESS-smoothed letter frequency distribution for that century. Gray lines represent the character frequency trends of the rest of the centuries for comparison.}
\label{fig:7} 
\end{figure}

\begin{figure}[th!]
    \centering
    \includegraphics[width=8cm, height=7cm, trim=6 7 4 26, clip]{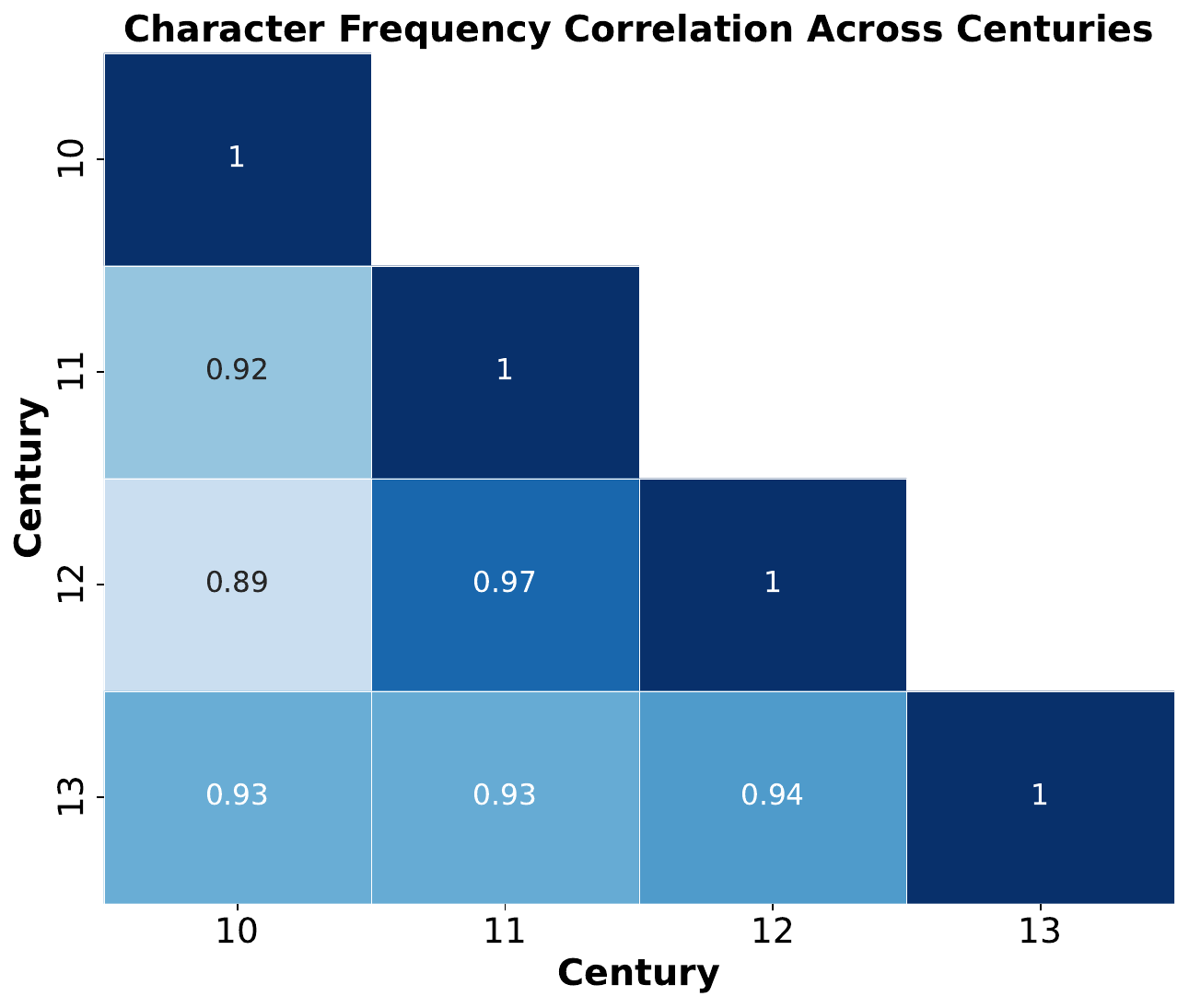}
    \vspace{-3mm}
\caption{Pearson correlation heatmap of the character frequency distribution across centuries in our synthetic dataset. Higher values (darker shades) indicate stronger similarity in character usage, while lower values (lighter shades) suggest greater divergence.}
\label{fig:8} 
\end{figure}

To further examine these variations, Figure~\ref{fig:7} visualizes the fitted frequency distributions per century using LOWESS smoothing. The black line represents the smoothed trend for a given century, while gray lines show distributions from the rest of the centuries for comparison. Lastly, Figure~\ref{fig:8} quantifies distributional similarity via a Pearson correlation heatmap. Adjacent centuries show strong correlations (e.g., $11$th–$12$th: $0.97$), while more distant ones exhibit lower values (e.g., $10$th–$12$th: $0.89$), indicating more pronounced shifts.

\section{EdAcc Dataset - Analysis}
\begin{figure}[th!]
    \centering
    \includegraphics[width=7.8cm, height=8cm, trim=5 5 6 7, clip]{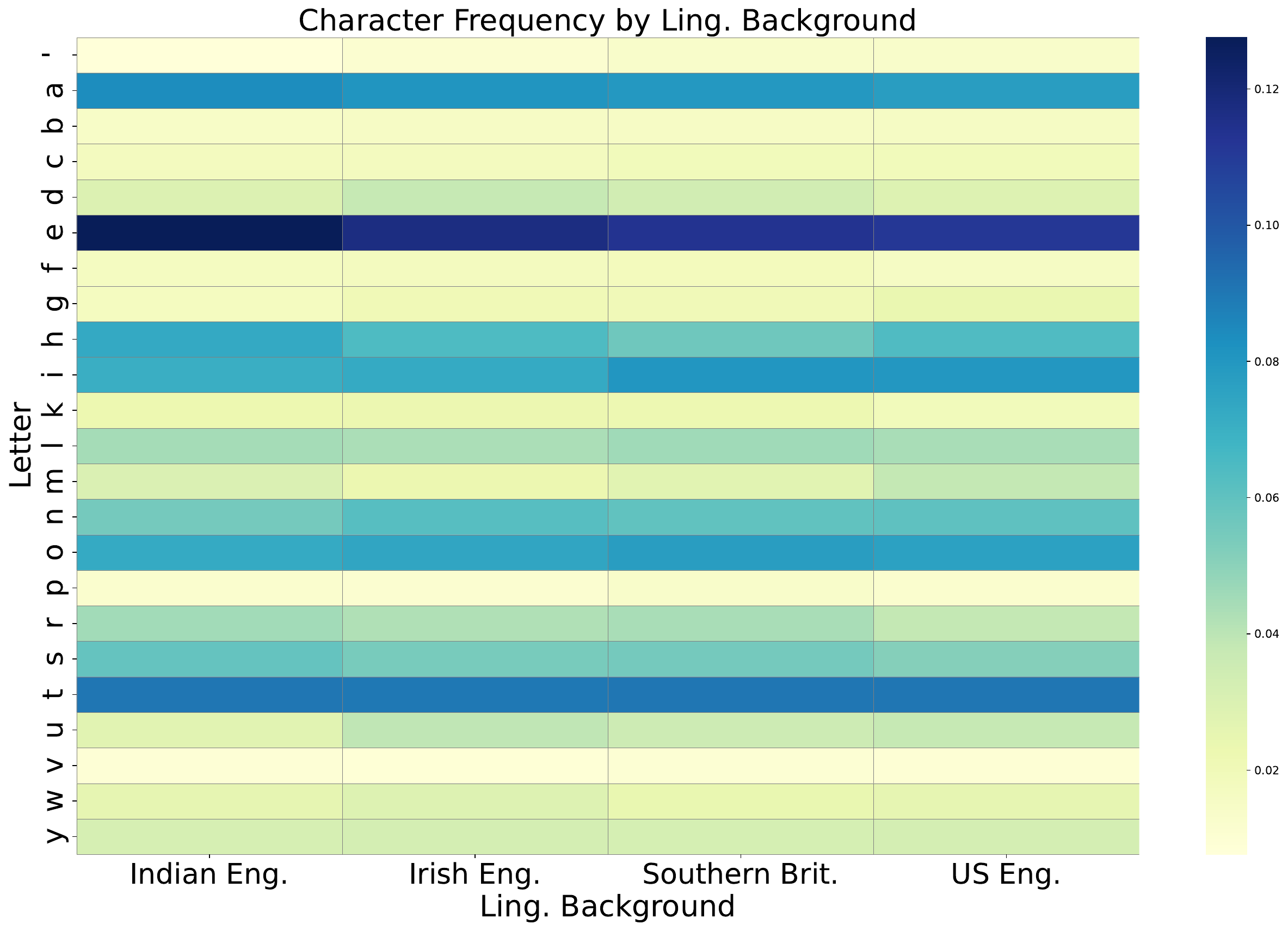}
    \vspace{-3mm}
\caption{Heatmap illustrating the relative frequency distribution of characters in EdAcc \cite{sanabria23edacc}.}
\label{fig:edacc} 
\end{figure}

\begin{figure}[th!]
    \centering
    \includegraphics[width=7.0cm, height=9.5cm, trim=5 5 6 7, clip]{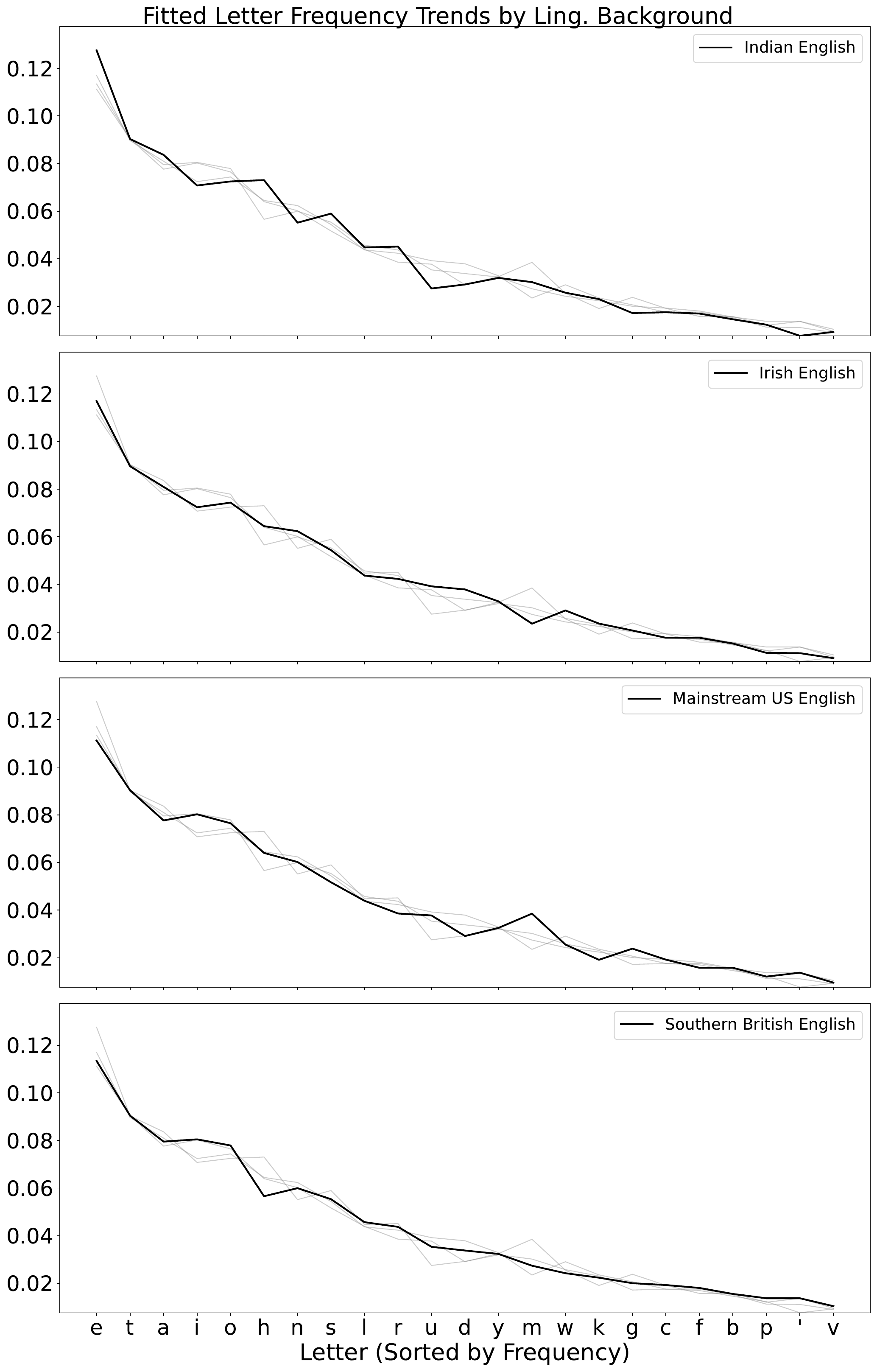}
    \vspace{-3mm}
\caption{Fitted character frequency trends across different linguistic backgrounds. Each subplot represents a specific background, with the black line indicating the LOWESS-smoothed letter frequency distribution for that background. Gray lines represent the character frequency trends of the rest of the backgrounds for comparison.}
\label{fig:10} 
\end{figure}

Similarly, we provide an analysis for our ASR dataset. Figure~\ref{fig:edacc} presents a heatmap visualizing the intra-dataset character frequency distributions among the four linguistic backgrounds, exploiting some subtle but noticeable patterns between them. Figure~\ref{fig:10} presents the fitted frequency distributions per linguistic background further enhancing the previously stated observations as the black lines deviate compared to the grey lines in each of the four plots.
\label{sec:appendix-edacc}

\section{Lipogram Generation - Extensive Results}
\label{sec:appendix-lipograms}

To further assess the impact of $\texttt{FADA}_\text{GD}$ beyond recognition tasks, we applied it in a controlled text generation setting focused on lipogram constraints. In this experiment, we tested whether $\texttt{FADA}_\text{GD}$ could steer an LLM (\texttt{Llama3.1}) to generate coherent passages while avoiding a specific character. The task involved generating short texts ($4$-$6$ sentences) on various topics while ensuring that a predetermined character never appeared. This provided a structured way to evaluate how effectively$\texttt{FADA}_\text{GD}$ can influence character distributions in an open-ended autoregressive setting.

We selected three different forbidden characters, namely ``A'', ``E'' and ``L''. In each case, we applied $\texttt{FADA}_\text{GD}$ during decoding by setting the forbidden character's target relative frequency to zero.To assess its impact, we compared two decoding configurations: standard beam search, where no constraints were applied, and \texttt{FADA-GD}, which introduced frequency-aware scoring to penalize sequences containing the forbidden character.

For each configuration, the model was prompted with $30$ different topics covering diverse subjects. The full list of topics is presented in Table \ref{tab:lipogram_topics}, while the prompt is presented in Figure \ref{fig:lipogram-prompt}. The generated text was evaluated based on the following metrics:

\begin{itemize}
    \item \textbf{Violation Rate}: The percentage of outputs where the forbidden character appeared at least once.
    \item \textbf{Perplexity}: A measure of fluency and coherence in the generated text.
    \item \textbf{Readability}: An assessment of how natural and comprehensible the generated text remains under the applied constraints. We calculated it  using the \textit{Flesch Reading Ease Score} \cite{reading-ease}, which evaluates how easy a text is to read. Higher scores indicate more readable text, with values above 60 considered easy to read, while lower scores (below 30) indicate more complex writing.
\end{itemize}

We present the numerical results in Table \ref{fig:lipogram_results}. The violation rate succesfully decreases across all three forbidden characters (``A'', ``E'', ``L''), with the most notable reduction observed for ``E'', where the percentage of outputs containing the forbidden character drops from 10.60\% to 8.66\%, a relative improvement of 18.3\%. This confirms that frequency-aware decoding successfully discourages character occurrences without the need for model retraining.

In terms of perplexity, which reflects the fluency and coherence of the generated text, the results show a marginal increase for A and E, while it slightly improves for L. This suggests that enforcing character-level constraints with$ \texttt{FADA}_{\text{GD}}$ does not significantly degrade text fluency, with differences remaining within acceptable bounds. Readability scores, measured using the Flesch Reading Ease score \cite{reading-ease}, exhibit improvements for ``A'' and ``L'', while remaining nearly unchanged for ``E''. Overall, these findings highlight $\texttt{FADA}_{\text{GD}}$ as an effective approach for guiding language model outputs at the character level while maintaining fluency and readability. 

\begin{table}[h!]
\centering
\begin{tabular}{|c|p{6.5cm}|}
\hline
\textbf{ID} & \textbf{Topic} \\ \hline
1  & A person who went to the post office. \\ \hline
2  & A dog at the park. \\ \hline
3  & A family that went to the beach. \\ \hline
4  & An elephant in the jungle. \\ \hline
5  & A person who went to the aquarium. \\ \hline
6  & A child who lost a toy at the mall. \\ \hline
7  & A bird flying over a quiet village. \\ \hline
8  & A person who got stuck in an elevator. \\ \hline
9  & A cat exploring a new garden. \\ \hline
10 & A teacher who helped a struggling student. \\ \hline
11 & A couple hiking in the mountains. \\ \hline
12 & A farmer working in the field. \\ \hline
13 & A boy who found a hidden cave. \\ \hline
14 & A woman who forgot her umbrella on a rainy day. \\ \hline
15 & A man fishing by the river. \\ \hline
16 & A team that won a soccer match. \\ \hline
17 & A child who met their favorite author. \\ \hline
18 & A chef preparing a special dish for a celebration. \\ \hline
19 & A group of friends camping under the stars. \\ \hline
20 & A traveler visiting an ancient temple. \\ \hline
21 & A robot learning how to read. \\ \hline
22 & A young girl who dreamed of flying. \\ \hline
23 & A scientist discovering a new species. \\ \hline
24 & A person who got lost in a museum. \\ \hline
25 & A dog who saved a child from danger. \\ \hline
26 & A boy who built a treehouse with his friends. \\ \hline
27 & A person who volunteered at an animal shelter. \\ \hline
28 & A musician performing on a busy street. \\ \hline
29 & A fisherman who caught a rare fish. \\ \hline
30 & A person planting trees in a park. \\ \hline
\end{tabular}
\caption{Topics used in the Lipogram Generation experiment.}
\label{tab:lipogram_topics}
\end{table}

\noindent The prompt used in our experiments was: 

\begin{figure}[ht]
\centering
\begin{minipage}{0.9\linewidth}
\small{
\begin{verbatim}
A lipogram is a form of writing where one
or more specific letters are deliberately
avoided throughout the text.
This constraint challenges writers to be
creative and resourceful, often resulting
in unique word choices and phrasing.
Lipograms can be fun and artistic
but also difficult, especially when
the omitted letter is common.
For example, avoiding the letter
'e'—the most frequently used letter
in English—requires significant
effort and skill.
Writers must find alternative ways
to express ideas without breaking
the constraint, which can lead
to inventive language.

TASK:  
Generate a short story
(maximum 150 words) about {topic}.
You have to follow a lipogram constraint.  
The lipogram constraint is that the letter
{forbidden_character} must not appear
anywhere in the text.
First, go over the task.
Then, explain how you avoided
using the letter {forbidden_character} . 

FORMAT:  
To ensure clarity and easy readability,
format your output into a JSON.
Use the following format:
{
    "Story": "<Generated_story>"
}

Text:
\end{verbatim}}
\end{minipage}
\caption{The prompt template used for our lipogram generation experiments}
\label{fig:lipogram-prompt}
\end{figure}

\normalsize

\begin{table*}[h!]
    \centering
    \begin{adjustbox}{max width=\textwidth}
    \begin{tabular}{|p{2.6cm}||p{0.95cm}|p{0.95cm}|p{0.95cm}||p{0.95cm}|p{0.95cm}|p{0.95cm}||p{0.95cm}|p{0.95cm}|p{0.95cm}|}
        \hline
         \multicolumn{10}{|c|}{\textbf{\texttt{Lipogram Generation}}}\\
        \hline
        \multicolumn{1}{|c||}{\multirow{2}{*}{}} & \multicolumn{3}{c||}{\small\textbf{Violation Rate} $\downarrow$} & \multicolumn{3}{c||}{\small\textbf{Perplexity} $\downarrow$}  & \multicolumn{3}{c|}{\small\textbf{Readability} $\uparrow$} \\
        \cline{2-10}
        & \centering\small \textit{A} & \centering\small \textit{E} & \centering\small \textit{L} & \centering\small \textit{A} & \centering\small \textit{E} & \small \textit{L} & \centering\small \textit{A} & \centering\small \textit{E} & \small \textit{L} \\
        \hline
        \small\centering \textbf{Llama3.1} & \small \centering 6.48 & \small \centering 10.60 & \small \centering 3.02 & \small \centering \textbf{3.64} & \small \centering \textbf{1.93} & \small 2.24 & \small \centering 72.40 & \small \centering \textbf{78.97} & \small 79.20\\
        \hline
        \small\centering \textbf{Llama3.1 + $\texttt{FADA}_\text{GD}$} & \small \centering \textbf{5.73} & \small \centering \textbf{8.66} & \small \centering \textbf{2.50} & \small \centering 3.81 & \small \centering 1.98 & \small \textbf{2.03} & \small \centering \textbf{75.61} & \small \centering 78.11 & \small \textbf{79.37} \\
        \hline
    \end{tabular}
    \end{adjustbox}
    \\
    \caption{The performance of $\texttt{FADA}_\text{GD}$ framework in the lipogram generation task.}
    \label{fig:lipogram_results}
\end{table*}

\section{Per Century/Region Results}
\label{sec:appendix-per-century}

To further assess the impact of \texttt{FADA} across different temporal and linguistic variations, we report per-century and per-region results in Tables~\ref{fig:htr_results_per_century_hpgtr}–\ref{fig:asr_results_per_century}. These results allow us to examine how frequency-aware alignment affects model performance in subsets characterized by distinct character distributions.

For HTR, the results on the \texttt{HPGTR} dataset reveal that \texttt{FADA} provides substantial benefits in centuries where character distributions deviate most from the overall dataset trend. This effect is particularly noticeable in later centuries, where orthographic variations and evolving linguistic conventions introduce additional challenges for recognition models. Similarly, for the \texttt{CATMuS} dataset, where distributional shifts are more subtle, \texttt{FADA} still contributes to performance improvements, particularly in centuries with greater deviations from the dataset-wide character frequencies. The \texttt{Synthetic} dataset further reinforces these observations, as the controlled frequency shifts across different subsets allow us to systematically evaluate the method’s effectiveness in handling predefined character distribution discrepancies. Across all cases, the most substantial improvements are observed when both training-time and inference-time alignment are applied together, highlighting their complementary nature.

For ASR, the results from the \texttt{EdAcc} dataset demonstrate that \texttt{FADA} adapts well to different linguistic backgrounds, helping the model better handle accent-based variations that lead to systematic shifts in character usage patterns. While training-time alignment consistently enhances performance across most linguistic categories, the impact of inference-time alignment varies. In some cases, guided decoding further refines predictions, while in others, its effect is more limited, suggesting that the degree of benefit depends on the severity of distributional shifts and the alignment between training data and inference-time distributions.

\begin{table*}[h!]
    \centering
    \begin{adjustbox}{max width=\textwidth}
    \begin{tabular}{|p{2cm}||p{1.45cm}|p{1.45cm}|p{1.45cm}|p{1.45cm}|p{1.45cm}|p{1.45cm}|p{1.45cm}|}
        \hline
         \multicolumn{8}{|c|}{\textbf{\texttt{HPGTR Dataset} - Per Century Results (CER)}} \\
        \hline
        \multicolumn{1}{|c||}{\multirow{4}{*}{}} & \multicolumn{7}{c|}{\small\textbf{TrOCR-base \cite{trocr}}}\\
        \cline{2-8}
        & \centering\small \textit{10} & \centering\small \textit{11} & \centering\small \textit{12} & \centering\small \textit{13} & \centering\small \textit{14} & \centering\small \textit{15} & \small \textit{16} \\
        \hline
        \small\centering \textbf{\texttt{Standard FT}} & \small \centering 8.99 & \small \centering 15.31 & \small \centering 27.18 & \small \centering 32.47 & \small \centering 50.06 & \small \centering 32.37 & \small 43.78\\
        \hline
        \small\centering \textbf{$\texttt{FADA}_\text{GD}$} & \small \centering 8.44 & \small \centering 14.74 & \small \centering 27.27 & \small \centering 30.76 & \small \centering 49.34 & \small \centering 31.61 & \small 46.15 \\
        \hline
        \small\centering \textbf{$\texttt{FADA}_\text{TR}$} & \small \centering 8.07 & \small \centering 12.64 & \small \centering 25.48 & \small \centering \textbf{27.98} & \small \centering 46.57 & \small \centering 27.32 & \small 42.09 \\
        \hline
        \small\centering \textbf{$\texttt{FADA}_\text{TR-GD}$} & \small \centering \textbf{8.04} & \small \centering \textbf{12.07} & \small \centering \textbf{25.10} & \small \centering 28.11 & \small \centering \textbf{45.82} & \small \centering \textbf{27.15} & \small \textbf{41.36} \\
        \hline
    \end{tabular}
    \end{adjustbox}
    \\
    \centering
    \begin{adjustbox}{max width=\textwidth}
    \begin{tabular}{|p{2cm}||p{1.45cm}|p{1.45cm}|p{1.45cm}|p{1.45cm}|p{1.45cm}|p{1.45cm}|p{1.45cm}|}
        \hline
        \multicolumn{1}{|c||}{\multirow{4}{*}{}} & \multicolumn{7}{c|}{\small\textbf{C-RNN \cite{crnn}}}\\
        \cline{2-8}
        & \centering\small \textit{10} & \centering\small \textit{11} & \centering\small \textit{12} & \centering\small \textit{13} & \centering\small \textit{14} & \centering\small \textit{15} & \small \textit{16} \\
        \hline
        \small\centering \textbf{\texttt{Standard FT}} & \small \centering 10.26 & \small \centering 16.11 & \small \centering 27.19 & \small \centering 38.78 & \small \centering 37.78  & \small \centering 31.41 & \small 38.54\\
        \hline
        \small\centering \textbf{$\texttt{FADA}_\text{GD}$} & \small \centering 10.20 & \small \centering 16.07 & \small \centering 27.08 & \small \centering 38.14 & \small \centering 37.11 & \small \centering 30.55 & \small 38.51 \\
        \hline
        \small\centering \textbf{$\texttt{FADA}_\text{TR}$} & \small \centering 10.13 & \small \centering 15.88 & \small \centering 26.31 & \small \centering 37.84 & \small \centering \textbf{35.75} & \small \centering \textbf{29.54} & \small 38.76 \\
        \hline
        \small\centering \textbf{$\texttt{FADA}_\text{TR-GD}$} & \small \centering \textbf{10.05} & \small \centering \textbf{15.77} & \small \centering \textbf{26.27} & \small \centering \textbf{37.79} & \small \centering 35.77 & \small \centering 29.59 & \small \textbf{38.47} \\
        \hline
    \end{tabular}
    \end{adjustbox}
    \\
    \caption{The per-century performance (in terms of CER) of TrOCR-base \cite{trocr} and C-RNN \cite{crnn} on the HPGTR dataset \cite{platanou-etal-2022-handwritten}. \texttt{Baseline} denotes the vanilla model, $\texttt{FADA}_{\text{GD}}$ and $\texttt{FADA}_{\text{TR}}$ denote the proposed inference-time and training-time alignment methods respectively, while $\texttt{FADA}_{\text{TR-GD}}$ indicates the combination of them.}
    \label{fig:htr_results_per_century_hpgtr}
\end{table*}

\begin{table}[h!]
    \centering
    \begin{adjustbox}{max width=\textwidth}
    \begin{tabular}{|p{1.4cm}||p{1.05cm}|p{1.05cm}|p{1.05cm}|p{1.05cm}|}
        \hline
         \multicolumn{5}{|c|}{\small\textbf{\texttt{CATMuS Dataset} - Per Century Results (CER)}} \\
        \hline
        \multicolumn{1}{|c||}{\multirow{4}{*}{}} & \multicolumn{4}{c|}{\small\textbf{TrOCR-base \cite{trocr}}}\\
        \cline{2-5}
        & \centering\small \textit{13} & \centering\small \textit{14} & \centering\small \textit{15} & \small \textit{16} \\
        \hline
        \small\centering \textbf{\texttt{Standard FT}} & \small \centering 12.28 & \small \centering \textbf{12.07} & \small \centering 9.47 & \small 5.76\\
        \hline
        \small\centering \textbf{$\texttt{FADA}_\text{GD}$} & \small \centering 12.11 & \small \centering 12.16 & \small \centering 9.46 & \small 5.79 \\
        \hline
        \small\centering \textbf{$\texttt{FADA}_\text{TR}$} & \small \centering \textbf{11.95} & \small \centering 12.16 & \small \centering 9.42 & \small 4.94 \\
        \hline
        \small\centering \textbf{$\texttt{FADA}_\text{TR-GD}$} & \small \centering 12.10 & \small \centering 12.17 & \small \centering \textbf{9.18} & \small \textbf{4.90} \\
        \hline
    \end{tabular}
    \end{adjustbox}
    \\
    \centering
    \begin{adjustbox}{max width=\textwidth}
    \begin{tabular}{|p{1.4cm}||p{1.05cm}|p{1.05cm}|p{1.05cm}|p{1.05cm}|}
        \hline
        \multicolumn{1}{|c||}{\multirow{4}{*}{}} & \multicolumn{4}{c|}{\small\textbf{C-RNN \cite{crnn}}}\\
        \cline{2-5}
        & \centering\small \textit{13} & \centering\small \textit{14} & \centering\small \textit{15} & \small \textit{16} \\
        \hline
        \small\centering \textbf{\texttt{Standard FT}} & \small \centering 20.34 & \small \centering 23.01 & \small \centering 15.79 & \small 8.92\\
        \hline
        \small\centering \textbf{$\texttt{FADA}_\text{GD}$} & \small \centering 20.11 & \small \centering 22.89 & \small \centering 15.77 & \small 8.73 \\
        \hline
        \small\centering \textbf{$\texttt{FADA}_\text{TR}$} & \small \centering 19.87 & \small \centering 21.36 & \small \centering 15.89 & \small 8.67 \\
        \hline
        \small\centering \textbf{$\texttt{FADA}_\text{TR-GD}$} & \small \centering \textbf{19.78} & \small \centering \textbf{21.20} & \small \centering \textbf{15.64} & \small \textbf{8.65} \\
        \hline
    \end{tabular}
    \end{adjustbox}
    \\
    \caption{The per-century performance (in terms of CER) of TrOCR-base \cite{trocr} and C-RNN \cite{crnn} on the CATMuS dataset \cite{catmus}. \texttt{Standard FT} denotes the fine-tuned backbone model, $\texttt{FADA}_{\text{GD}}$ and $\texttt{FADA}_{\text{TR}}$ denote the proposed inference-time and training-time alignment methods respectively, while $\texttt{FADA}_{\text{TR-GD}}$ indicates the combination of them.}
    \label{fig:htr_results_per_century_catmus}
\end{table}

\begin{table}[h!]
    \centering
    \begin{adjustbox}{max width=\textwidth}
    \begin{tabular}{|p{1.4cm}||p{1.05cm}|p{1.05cm}|p{1.05cm}|p{1.05cm}|}
        \hline
         \multicolumn{5}{|c|}{\small\textbf{\texttt{Synthetic Dataset} - Per Category Results (CER)}} \\
        \hline
        \multicolumn{1}{|c||}{\multirow{4}{*}{}} & \multicolumn{4}{c|}{\small\textbf{TrOCR-base \cite{trocr}}}\\
        \cline{2-5}
        & \centering\small \textit{GPT-4} & \centering\small \textit{Llama} & \centering\small \textit{Gemini} & \small \textit{Claude} \\
        \hline
        \small\centering \textbf{\texttt{Standard FT}} & \small \centering 3.96 & \small \centering 9.83 & \small \centering 5.38 & \small 4.43\\
        \hline
        \small\centering \textbf{$\texttt{FADA}_\text{GD}$} & \small \centering 20.21 & \small \centering 22.78 & \small \centering 5.23 & \small 4.39 \\
        \hline
        \small\centering \textbf{$\texttt{FADA}_\text{TR}$} & \small \centering \textbf{2.96} & \small \centering 8.87 & \small \centering 5.05 & \small 4.27 \\
        \hline
        \small\centering \textbf{$\texttt{FADA}_\text{TR-GD}$} & \small \centering 3.23 & \small \centering \textbf{8.19} & \small \centering \textbf{4.65} & \small \textbf{3.81} \\
        \hline
    \end{tabular}
    \end{adjustbox}
    \\
    \centering
    \begin{adjustbox}{max width=\textwidth}
    \begin{tabular}{|p{1.4cm}||p{1.05cm}|p{1.05cm}|p{1.05cm}|p{1.05cm}|}
        \hline
        \multicolumn{1}{|c||}{\multirow{4}{*}{}} & \multicolumn{4}{c|}{\small\textbf{C-RNN \cite{crnn}}}\\
        \cline{2-5}
        & \centering\small \textit{GPT-4} & \centering\small \textit{Llama} & \centering\small \textit{Gemini} & \small \textit{Claude} \\
        \hline
        \small\centering \textbf{\texttt{Standard FT}} & \small \centering 9.29 & \small \centering 14.77 & \small \centering 11.32 & \small 8.97\\
        \hline
        \small\centering \textbf{$\texttt{FADA}_\text{GD}$} & \small \centering 9.17 & \small \centering 13.29 & \small \centering 10.88 & \small 8.45 \\
        \hline
        \small\centering \textbf{$\texttt{FADA}_\text{TR}$} & \small \centering 8.07 & \small \centering 11.67 & \small \centering 9.16 & \small 5.41 \\
        \hline
        \small\centering \textbf{$\texttt{FADA}_\text{TR-GD}$} & \small \centering \textbf{8.03} & \small \centering \textbf{11.48} & \small \centering \textbf{9.12} & \small \textbf{5.21} \\
        \hline
    \end{tabular}
    \end{adjustbox}
    \\
    \caption{The per-century performance (in terms of CER) of TrOCR-base \cite{trocr} and C-RNN \cite{crnn} on our synthetic dataset. \texttt{Standard FT} denotes the fine-tuned backbone model, $\texttt{FADA}_{\text{GD}}$ and $\texttt{FADA}_{\text{TR}}$ denote the proposed inference-time and training-time alignment methods respectively, while $\texttt{FADA}_{\text{TR-GD}}$ indicates the combination of them.}
    \label{fig:htr_results_per_century_synth}
\end{table}

\begin{table}[h!]
    \centering
    \begin{adjustbox}{max width=\textwidth}
    \begin{tabular}{|p{1.4cm}||p{1.05cm}|p{1.05cm}|p{1.05cm}|p{1.05cm}|}
        \hline
         \multicolumn{5}{|c|}{\small\textbf{\texttt{Synthetic Dataset} - Per Category Results (CER)}} \\
        \hline
        \multicolumn{1}{|c||}{\multirow{4}{*}{}} & \multicolumn{4}{c|}{\small\textbf{Whisper-small \cite{whisper}}}\\
        \cline{2-5}
        & \centering\small \textit{Southern British English} & \centering\small \textit{Indian English} & \centering\small \textit{Irish English} & \hspace{0.3cm}\small \textit{US English} \\
        \hline
        \small\centering \textbf{\texttt{Standard FT}} & \small \centering 37.23 & \small \centering 25.85 & \small \centering 28.37 & \small 28.70\\
        \hline
        \small\centering \textbf{$\texttt{FADA}_\text{GD}$} & \small \centering 36.26 & \small \centering 25.76 & \small \centering 29.51 & \small 28.38 \\
        \hline
        \small\centering \textbf{$\texttt{FADA}_\text{TR}$} & \small \centering 35.71 & \small \centering 25.93 & \small \centering 27.37 & \small 27.62 \\
        \hline
        \small\centering \textbf{$\texttt{FADA}_\text{TR-GD}$} & \small \centering 37.05 & \small \centering 26.17 & \small \centering 27.92 & \small 27.74 \\
        \hline
    \end{tabular}
    \end{adjustbox}
    \\
    \caption{The per linguistic background performance (in terms of CER) of Whisper-small on the EdAcc dataset \cite{sanabria23edacc}. \texttt{Standard FT} denotes the vanilla model, $\texttt{FADA}_{\text{GD}}$ and $\texttt{FADA}_{\text{TR}}$ denote the proposed inference-time and training-time alignment methods respectively, while $\texttt{FADA}_{\text{TR-GD}}$ indicates the combination of them.}
    \label{fig:asr_results_per_century}
\end{table}

\section{Tuning $\lambda$}\label{sec:appendix-tuning}
\label{sec:appendix-tuning}
To balance transcription accuracy and frequency alignment, we conducted a hyperparameter search over $\lambda$ values ranging from $0.05$ to $0.95$ in increments of $0.05$. Both training-time ($\texttt{FADA}_{\text{TR}}$) and inference-time ($\texttt{FADA}_{\text{GD}}$) alignment were evaluated on the validation set using CER, WER, and NED as metrics. While no single $\lambda$ value was optimal across all datasets and models, we observed that values in the $0.3$–$0.6$ range generally led to better overall performance.

\section{Ablation on beam size}
\label{sec:appendix-beam-ablation}
As an ablation study, we investigate the effect of the beam size in \texttt{FADA}. To this end, we compare the performance of the standard fine-tuned model with standard beam search against $\texttt{FADA}_\text{TR}$. We explore beam sizes of $1$ (greedy decoding), $5$ and $7$. For both datasets (HPGTR and Synthetic), increasing the beam size in the fine-tuned TrOCR model yields only marginal improvements, particularly in CER and WER. We present the results in Table~\ref{tab:beam-size-ablation}. While a larger beam (e.g., 5 or 7) generally reduces errors slightly compared to greedy decoding (beam size = 1), the gains are relatively small. This suggests that the standard beam search does not significantly improve recognition performance beyond a certain threshold.

\begin{table}[h!]
\centering
\small
\setlength{\tabcolsep}{3.6pt} 
\renewcommand{\arraystretch}{1.2} 

\begin{tabular}{|l|ccc|ccc|}
\hline
\multicolumn{7}{|c|}{\textbf{Beam Size Ablation Results}} \\
\hline
\multirow{2}{*}{\textbf{Model}} & \multicolumn{3}{c|}{\textbf{CER} $\downarrow$} & \multicolumn{3}{c|}{\textbf{WER} $\downarrow$} \\
\cline{2-7}
& \textit{1} & \textit{5} & \textit{7} & \textit{1} & \textit{5} & \textit{7} \\
\hline
\multicolumn{7}{|c|}{\textbf{HPGTR}} \\
\hline
Standard FT         & 29.01 & 26.92 & 26.87 & 76.25 & 73.82 & 73.31 \\
\texttt{FADA}\textsubscript{TR} & \textbf{25.33} & \textbf{25.16} & \textbf{24.73} & \textbf{70.41} & \textbf{70.47} & \textbf{69.98} \\
\hline
\multicolumn{7}{|c|}{\textbf{Synthetic Dataset}} \\
\hline
Standard FT         & 6.44 & 5.92 & 5.91 & 18.98 & 17.89 & 17.86 \\
\texttt{FADA}\textsubscript{TR} & \textbf{5.46} & \textbf{5.30} & \textbf{5.28} & \textbf{17.10} & \textbf{16.86} & \textbf{16.85} \\
\hline
\end{tabular}

\caption{\textbf{Beam size ablation on HPGTR and Synthetic datasets.} We report CER and WER for beam sizes 1, 5, and 7.}
\label{tab:beam-size-ablation}
\end{table}

When applying $\texttt{FADA}_\text{TR}$, the improvements are more substantial, particularly for beam size = $1$, where it consistently outperforms the fine-tuned model even at beam size = $7$. This demonstrates that the proposed training-time alignment enhances the model’s inherent ability to generate more accurate transcriptions without relying heavily on an extensive search space. Notably, even with beam size = $1$, $\texttt{FADA}_\text{TR}$ achieves lower CER and WER than the fine-tuned model with beam size = 7, highlighting the effectiveness of frequency-aware training. Moreover, $\texttt{FADA}\text{TR}$ also benefits from a moderate increase in beam size, but the relative improvements diminish as the beam size grows.

\end{document}